\documentclass[10pt,journal,compsoc]{IEEEtran}
\ifCLASSOPTIONcompsoc
  \usepackage[nocompress]{cite}
\else
  \usepackage{cite}
\fi

\ifCLASSINFOpdf
  \usepackage[pdftex]{graphicx}
\else
  \usepackage[dvips]{graphicx}
\fi

\usepackage{amsfonts,mathtools,amssymb,amsthm}
\usepackage[ruled]{algorithm2e}
\usepackage{algorithmic}
\usepackage{caption,subcaption}
\usepackage{booktabs,multirow}
\usepackage{array}

\usepackage{todonotes}

\newcommand{\digit}[1]{\ensuremath{\vcenter{\hbox{\includegraphics[height=9pt]{img/MNIST/#1.png}}}}}

\allowdisplaybreaks

\begin{document}
\title{GEDI: GEnerative and DIscriminative Training for Self-Supervised Learning}
%
%
%
%

\author{Emanuele Sansone,
        Robin Manhaeve
\IEEEcompsocitemizethanks{\IEEEcompsocthanksitem E. Sansone and R. Manhaeve are with the Department
of Computer Science, KU Leuven, Leuven,
Belgium, 3001.\protect\\
E-mail: emanuele.sansone@kuleuven.be
}
}

%
%


\IEEEtitleabstractindextext{%
\begin{abstract}
Self-supervised learning is a popular and powerful method for utilizing large amounts of unlabeled data, for which a wide variety of training objectives have been proposed in the literature.
In this study, we perform a Bayesian analysis of state-of-the-art self-supervised learning objectives and propose a unified formulation based on likelihood learning. Our analysis suggests a simple method for integrating self-supervised learning with generative models, allowing for the joint training of these two seemingly distinct approaches. We refer to this combined framework as GEDI, which stands for GEnerative and DIscriminative training. Additionally, we demonstrate an instantiation of the GEDI framework by integrating an energy-based model with a cluster-based self-supervised learning model. Through experiments on synthetic and real-world data, including SVHN, CIFAR10, and CIFAR100, we show that GEDI outperforms existing self-supervised learning strategies in terms of clustering performance by a wide margin. We also demonstrate that GEDI can be integrated into a neural-symbolic framework to address tasks in the small data regime, where it can use logical constraints to further improve clustering and classification performance.
\end{abstract}

}

\maketitle

\IEEEdisplaynontitleabstractindextext

%
\IEEEpeerreviewmaketitle

\section{Introduction}\label{sec:introduction}
\IEEEPARstart{S}{elf}-supervised learning (SSL) has achieved impressive results in recent years due to its ability to learn good representations from a large amount of unlabeled data. SSL solutions have been applied to a variety of tasks in a range of application domains, from natural language processing to computer vision.

There has been a lot of effort to analyze existing SSL approaches in order to identify fundamental principles and improve the design of future SSL algorithms. However, the design of these algorithms is often based on the size/capacity of neural networks and the amount of training data. In this work, we aim to challenge this common practice and propose a solution that can work with smaller neural networks and smaller amounts of data.

To achieve this goal, we first provide a Bayesian interpretation of recent SSL objectives, identifying the underlying probabilistic graphical models for the main families of SSL approaches. This analysis allows us to develop a general framework (called GEDI) that highlights a connection to generative models. Based on this framework, we propose a new solution that combines an energy-based model with a state-of-the-art cluster-based SSL model. We show that this unified solution can outperform the corresponding solutions trained in isolation by a large margin. In particular, we show an improvement of 15, 2, and 3 percentage points in terms of clustering performance over state-of-the-art baselines on the SVHN, CIFAR-10, and CIFAR-100 datasets, respectively. Additionally, we demonstrate that our solution can be easily integrated into a neural-symbolic framework like DeepProbLog~\cite{manhaeve2018deepproblog}, and provide a proof-of-concept application to solve a neural-symbolic task in the small data regime. Indeed, we are able to show that the proposed method 
can leverage the information offered by the constraint to further improve clustering and classification performance.
The proposed method also helps the neural-symbolic framework from mode-collapse onto trivial solutions.

The article is structured as follows: In Section 2, we provide a Bayesian interpretation of three classes of SSL approaches, namely contrastive, cluster-based and negative-free methods. In Section 3, we propose a general recipe (GEDI) to unite generative and SSL models, and provide an implementation of the general framework using energy-based and cluster-based SSL solutions. In Section 4, we review related work on SSL, including different objectives, theoretical results, and connections to energy-based models. In Section 5, we analyze and discuss the experiments. Finally, in Section 6, we discuss the limitations and future research directions for SSL.

\section{A Bayesian Interpretation of SSL}
\begin{figure*}
     \centering
     \begin{subfigure}[b]{0.24\textwidth}
         \centering \includegraphics[width=0.43\textwidth]{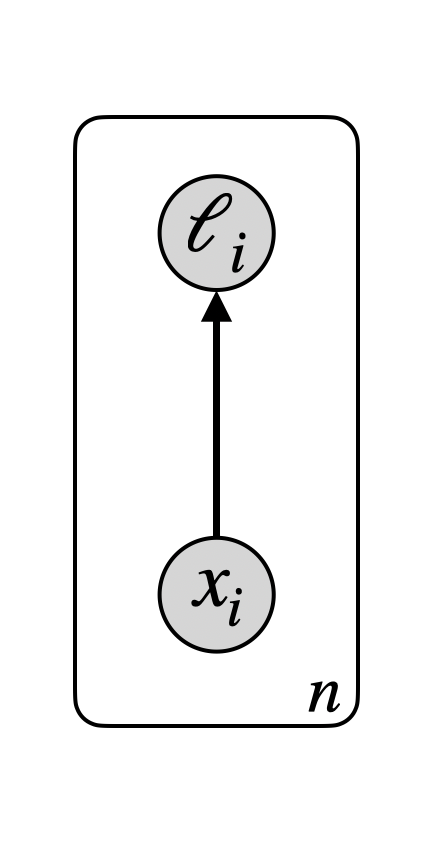}
         \caption{Contrastive (CT)}
     \end{subfigure}%
     \begin{subfigure}[b]{0.24\textwidth}
         \centering      \includegraphics[width=0.71\textwidth]{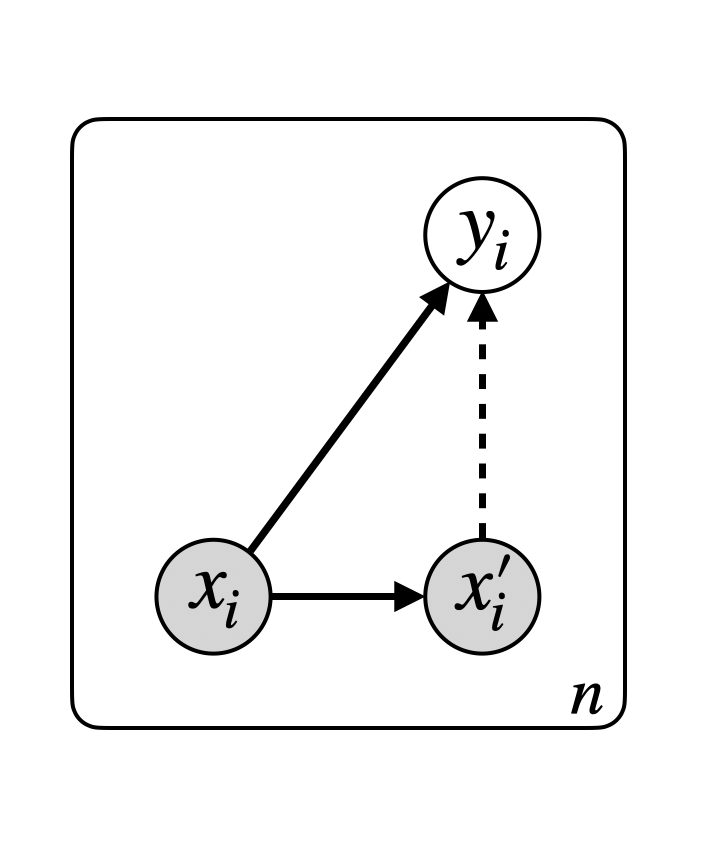}
         \caption{Discriminative (DI)}
     \end{subfigure}%
     \begin{subfigure}[b]{0.24\textwidth}
         \centering        \includegraphics[width=0.89\textwidth]{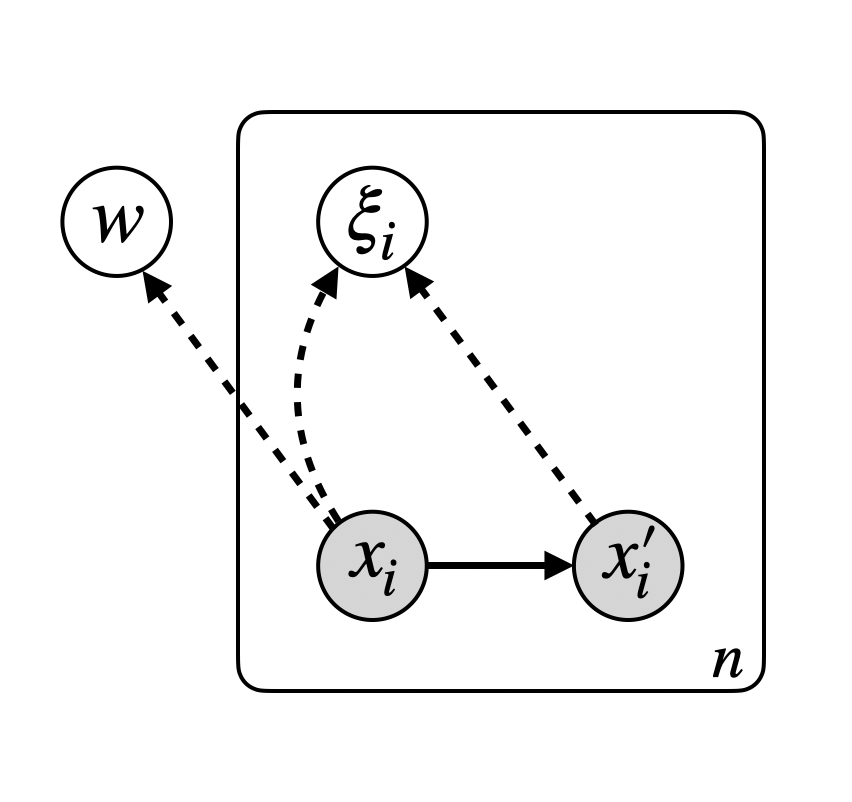}
         \caption{Negative-Free (NF)}
     \end{subfigure}%
     \begin{subfigure}[b]{0.24\textwidth}
         \centering       \includegraphics[width=1.15\textwidth]{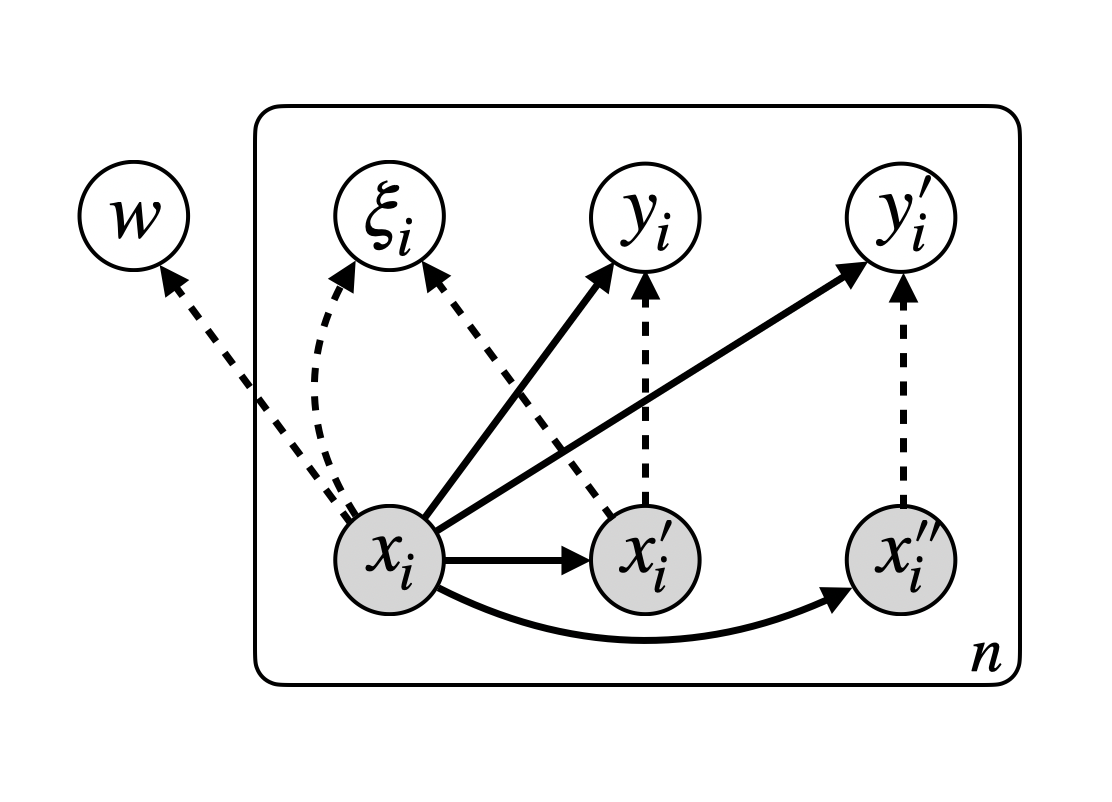}
         \caption{GEDI}
     \end{subfigure}%
     \caption{Probabilistic graphical models for the different classes of self-supervised learning approaches. White and grey nodes represent hidden 
 and observed vectors/variables, respectively. Solid arrows define the generative process, whereas dashed arrows identify auxiliary posterior densities/distributions.}
     \label{fig:graph}
\end{figure*}
Let us introduce the random quantities used throughout this section to analyze self-supervised learning approaches. We use (i) $x_i\in\Omega$, where $\Omega$ is a compact subset of $\mathbb{R}^d$, to identify observed data drawn independently from an unknown distribution $p(x)$, (ii) $x_i'\in\Omega$ to identify observed data drawn independently from a stochastic augmentation strategy $\mathcal{T}(x'|x)$ and (iii) $x_i''\in\Omega$ to identify data drawn independently from a generative model $p_\Psi(x'')$ (with parameters $\Psi$). We use $\xi_i\in\mathbb{R}^h$, $z_i\in\mathcal{S}^{h-1}$, where $\mathcal{S}^{h-1}$ is a $h-1$ dimensional unit hypersphere, and $w\in\mathbb{R}^l$, to identify a latent representation, a latent embedding and a global latent representation, respectively. Importantly, the latent representation is obtained through an encoding function $enc:\Omega\rightarrow\mathbb{R}^h$, whereas the latent embedding is obtained by an embedding function $g:\Omega\rightarrow\mathcal{S}^{h-1}$, defined as the composition of $proj:\mathbb{R}^h\rightarrow\mathcal{S}^{h-1}$ (a.k.a. projection head in the deep learning community) and $enc$, and the global representation summarizes the information common to all instances in a dataset (we are going to see its precise definition later on). Finally, we introduce $y_i,y_i'\in\{1,\dots,c\}$ to identify the latent labels over $c$ categories associated to input data and their augmented version, respectively. $\ell_i$ is a categorical variable defined on $n$ classes.  Notably, i is used to index different samples, viz. $i = \{1,\dots,n\}$

We distinguish self-supervised learning approaches according to three different classes: 1) contrastive, 2) cluster-based (or discriminative) and, 3) negative-free (or non-contrastive) methods. Fig.~\ref{fig:graph} shows the corresponding probabilistic graphical models for each of the self-supervised learning classes. From Fig.~\ref{fig:graph}, we can also see how the different random quantities, we have just introduced, are probabilistically related to each other.

We are now ready to give an interpretation of the different SSL classes.

\subsection{Contrastive SSL}\label{sec:contrastivessl}
Contrastive self-supervised learning can be interpreted in probabilistic terms using the graphical model in Fig.~\ref{fig:graph}(a). In particular, we can define the following conditional density:
\begin{equation}
p(\ell_i|x_i;\Theta)=\frac{e^{sim(g(x_{\ell_i}),g(x_i))/\tau}}{\sum_{j=1}^ne^{sim(g(x_j),g(x_i))/\tau}}
\label{eq:contrastive}
\end{equation}
where $sim$ is a similarity function, $\tau>0$ is temperature parameter used to calibrate the uncertainty for $p(\ell_i|x_i;\Theta)$ and $\Theta=\{\theta,\{x_i\}_i^n\}$ is the set of parameters, including the parameters of the embedding function and the observed data. The learning criterion can be obtained from the expected log-likelihood computed on the observed random quantities (and using the factorization provided by the graphical model), namely:
\begin{align}
    &\mathbb{E}_{p(x_{1:n},\ell_{1:n})}\big\{\log p(x_{1:n},\ell_{1:n};\Theta)\big\} \nonumber \\
    &\qquad= \mathbb{E}_{\prod_{j=1}^np(x_j)\delta(\ell_j-j)}\big\{\log \prod_{i=1}^np(x_i)p(\ell_i|x_i;\Theta)\big\} \nonumber\\
    &\qquad=\sum_{i=1}^n \mathbb{E}_{p(x_i)}\big\{\log p(x_i)\big\} \nonumber\\
    &\qquad\qquad+\mathbb{E}_{\prod_{j=1}^np(x_j)\delta(\ell_j-j)}\bigg\{\sum_{i=1}^n\log p(\ell_i|x_i;\Theta)\bigg\} \nonumber\\
    &\qquad= \underbrace{\sum_{i=1}^n \mathbb{E}_{p(x_i)}\big\{\log p(x_i)\big\}}_\text{Negative entropy term, $-H_p(x_{1:n})$} \nonumber\\
    &\qquad\qquad+ \underbrace{\mathbb{E}_{\prod_{j=1}^np(x_j)}\bigg\{\sum_{i=1}^n\log p(\ell_i=i|x_i;\Theta)\bigg\}}_\text{Conditional log-likelihood term $\mathcal{L}_{CT}(\Theta)$}
    \label{eq:contrastive_obj}
\end{align}
where $\delta$ in the second equality is a delta function and we use for instance $x_{1:n}$ as a compact way to express $x_1,\dots,x_n$. From Eq.~(\ref{eq:contrastive_obj}), we observe that the expected log-likelihood can be rewritten as the sum of two quantities, namely a negative entropy and a conditional log-likelihood terms. However, only the second addend in Eq.~(\ref{eq:contrastive_obj}) is relevant for maximization purposes over the parameters $\theta$. Importantly, we can now show that the conditional log-likelikood term in Eq.~(\ref{eq:contrastive_obj}), coupled with the definition provided in Eq.~(\ref{eq:contrastive}), corresponds to the notorious InfoNCE objective~\cite{oord2018representation}. To see this, let us recall InfoNCE:
\begin{align}
    \text{InfoNCE}\propto \mathbb{E}_{\prod_{j=1}^np(x_j,z_j)}\bigg\{\sum_{i=1}^n\log\frac{e^{f(x_i,x_i)}}{\sum_{k=1}^ne^{f(x_k,x_i)}}\bigg\} \nonumber
\end{align}
By choosing $p(x,z)=p(x)\delta(z-g(x))$ and $f(x,z)=sim(g(x),z)/\tau$ we recover the conditional log-likelihood term in Eq.~(\ref{eq:contrastive_obj}). Importantly, other contrastive objectives, such as CPC~\cite{henaff2020data}, SimCLR~\cite{chen2020simple}, ProtoCPC~\cite{lee2022prototypical}, KSCL~\cite{xu2022k} to name a few, can be obtained once we have a connection to InfoNCE. In the Supplementary Material, we provide an additional and alternative analysis of contrastive learning approaches (cf. Section A), we recall the connections with mutual information objectives (cf. Section B) and we show the derivation of InfoNCE (cf. Section C) and an example of corresponding surrogate objective based on ProtoCPC (cf. Section D).

\subsection{Discriminative/Cluster-Based SSL}
Cluster-based SSL can be interpreted in probabilistic terms using the graphical model in Fig.~\ref{fig:graph}(b). In particular, we can define the following conditional density:
\begin{equation}
p(y_i|x_i;\Theta)=\frac{e^{U_{:y_i}^TG_{:i}/\tau}}{\sum_{y}e^{U_{:y}^TG_{:i}/\tau}}
\label{eq:discriminative}
\end{equation}
where $U\in\mathbb{R}^{h\times c}$ is a matrix\footnote{We use subscripts to select rows and columns. For instance, $U_{:y}$ identify $y-$th column of matrix $U$.} of $c$ cluster centers, $G=[g(x_1),\dots,g(x_n)]$ is a matrix of embeddings of size $h\times n$ and $\Theta=\{\theta,U\}$ is the set of parameters, including the ones for the embedding function and the cluster centers.  The learning criterion can be obtained in a similar way to what we have done previously for contrastive methods. In particular, we have that
\begin{align}
&\mathbb{E}_{p(x_{1:n})}\{\log p(x_{1:n};\Theta)\}\nonumber\\
& =-H_p(x_{1:n})+ \mathbb{E}_{p(x_{1:n})\mathcal{T}(x_{1:n}'|x_{1:n})}\left\{\log\sum_{y_{1:n}}p(y_{1:n}|x_{1:n}';\Theta)\right\}\nonumber\\
&=-H_p(x_{1:n})+ \mathbb{E}_{p(x_{1:n})\mathcal{T}(x_{1:n}'|x_{1:n})}\Big\{\log\sum_{y_{1:n}}\frac{q(y_{1:n}|x_{1:n}')}{q(y_{1:n}|x_{1:n}')} \nonumber\\
&\qquad\qquad\qquad\qquad\qquad\qquad\qquad\qquad\qquad p(y_{1:n}|x_{1:n}';\Theta)\} \nonumber\\
&\geq -H_p(x_{1:n})- \mathbb{E}_{p(x_{1:n})\mathcal{T}(x_{1:n}'|x_{1:n})}\{KL(q(y_{1:n}|x_{1:n}')\|\nonumber\\
&\qquad\qquad\qquad\qquad\qquad\qquad\qquad\qquad\qquad p(y_{1:n}|x_{1:n}))\} \nonumber\\
&=\underbrace{\sum_{i=1}^n \mathbb{E}_{p(x_i)}\big\{\log p(x_i)\big\}}_\text{Negative entropy term, $-H_p(x_{1:n})$}\nonumber\\
&\quad+\underbrace{\sum_{i=1}^n\mathbb{E}_{p(x_i)\mathcal{T}(x_i'|x_i)}\{\mathbb{E}_{q(y_i|x_i')}\log p(y_i|x_i;\Theta) + H_q(y_i|x_i')\}}_\text{Discriminative term $\mathcal{L}_{DI}(\Theta;\mathcal{T})$}
\label{eq:discriminative_obj}
\end{align}
where $q(y_i|x_i')$ is an auxiliary distribution. Notably, maximizing the discriminative term is equivalent to minimize the $KL$ divergence between the two predictive distributions $p(y_i|x_i)$ and $q(y_i|x_i')$, thus learning to predict similar category for both sample $x_i$ and its augmented version $x_i'$, obtained through $\mathcal{T}$. Importantly, we can relate the criterion in Eq.~(\ref{eq:discriminative_obj}) to the objective 
used in optimal transport~\cite{cuturi2013sinkhorn}, by substituting Eq.~(\ref{eq:discriminative}) into Eq.~(\ref{eq:discriminative_obj}) and adopting a matrix format, namely:
\begin{align}
    \mathcal{L}_{DI}(\Theta;\mathcal{T}) &=\frac{1}{\tau}\bigg\{ \mathbb{E}_{p(x_{1:n})\mathcal{T}(x_i'|x_i)}\{Tr(QU^TG)\}&\nonumber\\
    &\qquad+\tau\mathbb{E}_{p(x_{1:n})\mathcal{T}(x_i'|x_i)}\{H_Q(y_{1:n}|x_{1:n}')\}\bigg\} 
    \label{eq:discriminative_obj2}
\end{align}
where $Q=[q(y_1|x_1'),\dots,q(y_n|x_n')]^T$ is a prediction matrix of size $n\times c$ and $Tr(A)$ is the trace of an arbitrary matrix $A$. Note that a naive maximization of $\mathcal{L}_{DI}(\Theta)$ can lead to obtain trivial solutions like 
the one corresponding to uniformative predictions, namely $q(y_i|x_i')=p_\gamma(y_i|x_i)=\text{Uniform}(\{1,\dots,c\})$ for all $i=1,\dots,n$. Fortunately, the problem can be avoided and solved exactly using the Sinkhorn-Knopp algorithm, which alternates between maximizing $\mathcal{L}_{DI}(\Theta)$ in Eq.~(\ref{eq:discriminative_obj2}) with respect to $Q$ and with respect to $\Theta$, respectively. This is indeed the procedure used in several cluster-based SSL approaches, like DeepCluster~\cite{caron2018deep} and SwAV~\cite{caron2020unsupervised}, to name a few.

\subsection{Non-contrastive/Negative-Free SSL}
We can provide a probabilistic interpretation also for negative-free SSL using the graphical model shown in Fig.~\ref{fig:graph}(c). Note that, for the sake of simplicity in the graph and in the following derivation, the latent variables ($w$ and all $\xi_i$) are considered independent of each other only for the generation process. In reality, one should consider an alternative but equivalent model, using a generating process including also the edges $x_i\rightarrow w$, $\xi_i\rightarrow x_i'$ and defining $p(w|x_i)=p(w)$ and $p(x_i'|x_i,\xi_i)=\mathcal{T}(x_i'|x_i)$ for all $i=1\dots,n$. Based on these considerations, we can define the prior and our auxiliary and inference densities for the model in the following way:
\begin{align}
    p(w) &= \mathcal{N}(w|0,I) \nonumber\\
    p(\xi_i) &= \mathcal{N}(\xi_i|0,I) \nonumber\\
    q(w|x_{1:n};\Theta) &=\mathcal{N}(w|0,\Sigma) \nonumber\\
    q(\xi_i|x_i,x_i';\Theta) &= \mathcal{N}(\xi_i|enc(x_i)-enc(x_i'),I)
    \label{eq:negativefree}
\end{align}
where $\mathcal{N}(\cdot|\mu,\Sigma)$ refers to a multivariate Gaussian density with mean $\mu$ and covariance $\Sigma$, $I$ is an identity matrix, $\Sigma=\sum_{i=1}^n(g(x_i)-\bar{g})(g(x_i)-\bar{g})^T+\beta I$, $\beta$ is positive scalar used to ensure the positive-definiteness of $\Sigma$, $\bar{g}=1/n\sum_{i=1}^ng(x_i)$ and $\Theta=\{\theta\}$.
Importantly, while $q(w|x_{1:n})$ in Eq.~(\ref{eq:negativefree}) is used to store the global statistical information of the data in the form of an unnormalized sample covariance, $q(\xi_i|x_i,x_i')$ is used to quantify the difference between a sample and its augmented version in terms of their latent representation.
Similarly to previous SSL classes and by reusing definitions in Eq.~(\ref{eq:negativefree}), we can devise the learning criterion in the following way:
\begin{align}
&\mathbb{E}_{p(x_{1:n})}\{\log p(x_{1:n};\Theta)\}\nonumber\\
&\geq -H_p(x_{1:n}) - \mathbb{E}_{p(x_{1:n})}\{KL(q(w|x_{1:n};\Theta)\|p(w))\} \nonumber\\
& \quad -\mathbb{E}_{p(x_{1:n})\mathcal{T}(x_{1:n}'|x_{1:n})}\{KL(q(\xi_{1:n}|x_{1:n},x_{1:n}';\Theta)\|p(\xi_{1:n}))\}\nonumber\\
&= -H_p(x_{1:n}) - \mathbb{E}_{p(x_{1:n})}\{KL(q(w|x_{1:n};\Theta)\|p(w))\} \nonumber\\
& \quad -\sum_{i=1}^n\mathbb{E}_{p(x_i)\mathcal{T}(x_i'|x_i)}\{KL(q(\xi_i|x_i,x_i';\Theta)\|p(\xi_i))\}\nonumber\\
&\propto\underbrace{\sum_{i=1}^n \mathbb{E}_{p(x_i)}\big\{\log p(x_i)\big\}}_\text{Negative entropy term, $-H_p(x_{1:n})$}\underbrace{-\mathbb{E}_{p(x_{1:n})}\left\{\frac{Tr(\Sigma)}{2}{-} \frac{\log|\Sigma|}{2}\right\}}_\text{Negative-free term, $\mathcal{L}_{NF}(\Theta)$ (first part)}\nonumber\\
& \qquad\underbrace{-\sum_{i=1}^n\mathbb{E}_{p(x_i)\mathcal{T}(x_i'|x_i)}\left\{\frac{dist(x_i,x_i')}{2}\right\}}_\text{Negative-free term, $\mathcal{L}_{NF}(\Theta)$ (second part)}
\label{eq:negativefree_obj}
\end{align}
where $dist(x,x')=\|enc(x)-enc(x')\|^2$ and $|A|$ computes the determinant of an arbitrary matrix $A$. Notably, the maximization of $\mathcal{L}_{NF}(\Theta)$ promotes both decorrelated embedding features, as the first two addends in $\mathcal{L}_{NF}(\Theta)$ (obtained from the first KL term in Eq.~(\ref{eq:negativefree_obj})) force $\Sigma$ to become an identity matrix, as well as representations that are invariant to data augmentations, thanks to the third addend in $\mathcal{L}_{NF}(\Theta)$. It is important to mention that $\mathcal{L}_{NF}(\Theta)$ in Eq.~(\ref{eq:negativefree_obj}) recovers two recent negative-free criteria, namely CorInfoMax~\cite{ozsoy2022self} and MEC~\cite{liu2022self}. We can also relate $\mathcal{L}_{NF}(\Theta)$ to other existing negative-free approaches, including Barlow Twins~\cite{zbontar2021barlow}, VicReg~\cite{bardes2022vicreg,bardes2022vicregl} and W-MSE~\cite{ermolov2021whitening} (please refer to Section E in the Supplementary Material for further details).


\section{Unifying Generative and SSL Models: A General Recipe (GEDI)}
In all three classes of SSL approaches (see Eqs.~(\ref{eq:contrastive_obj}),(\ref{eq:discriminative_obj}) and (\ref{eq:negativefree_obj})), the expected data log-likelihood can be lower bounded by the sum of two contribution terms, namely a negative entropy $-H_p(x_{1:n})$ and a conditional log-likelihood term, chosen from $\mathcal{L}_{CT}(\Theta),\mathcal{L}_{DI}(\Theta;\mathcal{T})$ and $\mathcal{L}_{NF}(\Theta)$. A connection to generative models emerges by additionally lower bounding the negative entropy term, namely:
\begin{align}
    -H_p(x_{1:n})&=\mathbb{E}_{p(x_{1:n})}\{\log p(x_{1:n})\} \nonumber\\
    &=\sum_{i=1}^n\mathbb{E}_{p(x_i)}\{\log p(x_i)\} \nonumber\\
    &=\sum_{i=1}^n\left[\mathbb{E}_{p(x_i)}\{\log p_\Psi(x_i)\}+KL(p(x_i)\|p_\Psi(x_i))\right] \nonumber\\
    &\geq \underbrace{\sum_{i=1}^n\mathbb{E}_{p(x_i)}\{\log p_\Psi(x_i)\}}_\text{$-CE(p,p_\Psi)$}
    \label{eq:generative_obj}
\end{align}
where $p_\Psi(x)$ is a generative model parameterized by $\Psi$. Notably, the relation in~(\ref{eq:generative_obj}) can be substituted in any of the objectives previously derived for the different SSL classes, thus allowing to jointly learn both generative and SSL models. This leads to a new GEnerative and DIscriminative family of models, which we call GEDI. Importantly, any kind of likelihood-based generative model (for instance variational autoencoders, normalizing flows, autoregressive or energy-based models) can be considered in GEDI. In this work, we argue that much can be gained by leveraging the GEDI integration. Notably, there has been a recent work EBCLR~\cite{kim2022energy} integrating energy-based  models with contrastive SSL approaches. Here, we show that EBCLR represents one possible instantiation of GEDI. For instance, let us consider contrastive SSL and observe that the conditional density in Eq.~(\ref{eq:contrastive}) can be decomposed into a joint and a marginal densities (similarly to what is done in~\cite{grathwohl2020your}):
\begin{align}
    p(\ell,x;\Theta) &= \frac{e^{sim(g(x_{\ell}),g(x))/\tau}}{\Gamma(\Theta)} \nonumber\\
    p(x;\Theta) &= \frac{\sum_{j=1}^ne^{sim(g(x_{\ell_j}),g(x))/\tau}}{\Gamma(\Theta)} \nonumber\\
    &= \frac{e^{-\underbrace{\left(-\log\sum_{j=1}^ne^{sim(g(x_{\ell_j}),g(x))/\tau}\right)}_\text{$E(x;\Theta)$}}}{\Gamma(\Theta)}
    \label{eq:generative_ebm}
\end{align}
where $E(x,\Theta)$ defines the energy score of the marginal density. Now, by choosing $p_{\Psi}(x)=p(x;\Theta)$ and $sim(z,z')=-\|z-z'\|^2$ in Eq.~(\ref{eq:generative_ebm}), one recovers the exact formulation of EBCLR~\cite{kim2022energy}.

In the subsequent sections, we are going to provide a new GEDI instantiation, by specifically integrating energy-based models with cluster-based SSL approaches.
We are also going to show the benefits arising from such integration.

\subsection{GEDI Instantiation: Energy-based and cluster-based SSL}
\begin{figure}
    \centering    \includegraphics[width=0.8\linewidth]{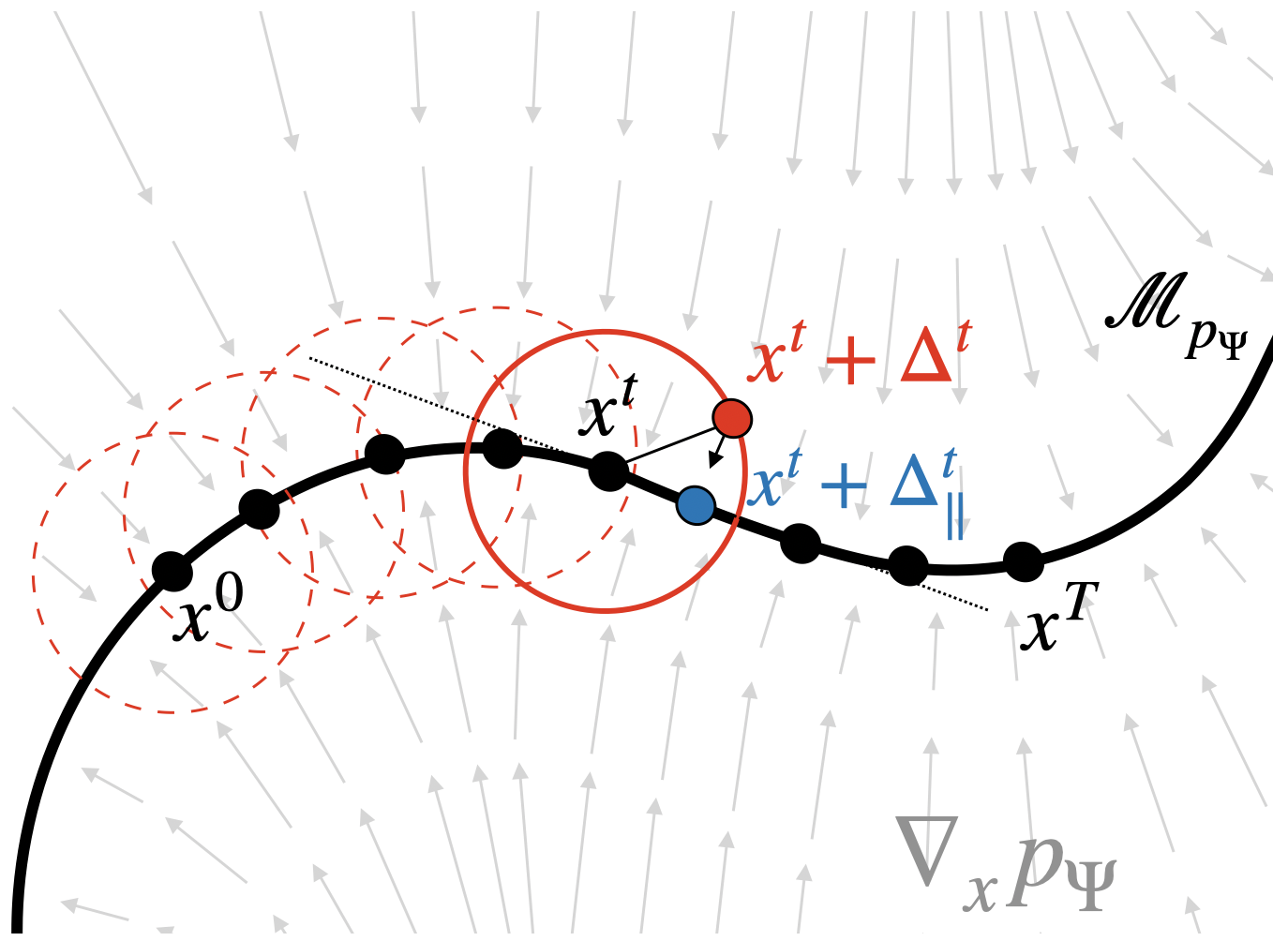}
    \caption{Visualization of the data augmentation strategy exploiting the information about the manifold structure $\mathcal{M}_{p_\Psi}$ and the vector field $\nabla_xp_\Psi$ induced by the energy-based model $p_\Psi$. A local perturbation $\Delta^t$ of a point $x^t$ is projected back onto the tangent plane of the manifold by using the gradient information. The strategy is applied iteratively starting from $x_i\equiv x^0$ up to $x_i''\equiv x^T$.}
     \label{fig:data_aug}
\end{figure}
For the GEDI instantiation, we consider the graphical model in Fig.~\ref{fig:graph}(d) and we derive the lower bound on the expected log-likelihood (exploiting also the bound in Eq.~(\ref{eq:generative_obj})) in a similar manner to what we have done for the three SSL classes, namely:
\begin{align}
    &\mathbb{E}_{p(x_{1:n})}\{\log p(x_{1:n};\Theta)\} \geq \underbrace{-CE(p,p_\Psi)}_\text{Generative term} \nonumber\\
    &\qquad
    + \underbrace{\mathcal{L}_{NF}(\Theta)+\mathcal{L}_{DI}(\Theta,\mathcal{T})+\mathcal{L}_{DI}(\Theta;p_\Psi)}_\text{Discriminative terms}
    \label{eq:gedi_obj}
\end{align}
where $p_\Psi(x)=\frac{e^{u^Tenc(x)}}{\Gamma(\Psi)}$ with $\Psi=\{\theta,u\}$, $\Theta=\{\theta,U\}$, both negative-free and cluster-based losses are included in the objective (an ablation study in the experimental evaluation will clarify the contribution of each loss term) and additionally we consider a cluster-based loss $\mathcal{L}_{DI}(\Theta;p_\Psi)$, which is computed using samples generated through a newly proposed procedure (called Data Augmentation based on Manifold structure) based on generative model $p_\Psi$.

\textbf{Data Augmentation based on Manifold structure (DAM).} The routine uses the energy-based model to generate samples by "walking" on the approximated data manifold and to assign same labels to these samples through the discriminative loss $\mathcal{L}_{DI}(\Theta,p_\Psi)$, thus enforcing the manifold assumption, commonly used in semi-supervised learning~\cite{chapelle2010semi}. DAM consists of an iterative procedure starting from the original training data $x_{1:n}$, drawn independently from $p$, and generating new samples $x_{1:n}''$ along the approximated data manifold induced by $p_\Psi$. At each iteration $t$, the algorithm performs two main operations: Firstly, it locally perturbs a sample $x^t$ by randomly choosing a vector $\Delta^t$ on a ball of arbitrarily small radius $\epsilon>0$, viz. $B_\epsilon$, and secondly, it projects the perturbed sample $x^t+\Delta^t$ back onto the tangent plane of the approximated data manifold $\mathcal{M}_{p_\Psi}$ using the following update rule:
\begin{algorithm}[t]
 \caption{Data Augmentation based on Manifold structure (DAM).}
 \label{alg:dam}
    \textbf{Input:} $x_{1:n}, p_\Psi, \epsilon, T$\;
    \textbf{Output:} $x_{1:n}''$\;
    Sample $\Delta^0$ uniformly at random from $B_\epsilon$\;
    \textbf{For} $t=0,\dots,T$\;
    \qquad Evaluate $\nabla_xp_\Psi(x_i^t+\Delta^t)$ for all $i=1,\dots,n$\;
    \qquad Update $x_i^t$ using Eq.~(\ref{eq:update_rule}) for all $i=1,\dots,n$\;
    \qquad $\Delta^t\leftarrow \Delta^0$\;
    $x_i''\leftarrow x_i^T$ for all $i=1,\dots,n$\;
    \textbf{Return} $x_{1:n}''$\;
\end{algorithm}
\begin{align}
    x^t\leftarrow x^t + \underbrace{\Delta^t -\left(\frac{\nabla_xp_\Psi(x^t+\Delta^t)^T\Delta^t}{\|\nabla_xp_\Psi(x^t+\Delta^t)\|^2}\right)\nabla_xp_\Psi(x^t+\Delta^t)}_\text{$\Delta_\parallel^t$}
    \label{eq:update_rule}
\end{align}
A visual interpretation as well as a complete description of the strategy are provided in Figure~\ref{fig:data_aug} and Algorithm~\ref{alg:dam}, respectively.

\textbf{Learning a GEDI model.} As we will see in the experiments, the data augmentation strategy proves effective when the energy-based model approximates well the unknown data density $p(x)$. Consequently, we opt to train our GEDI instantiation based on a two-step procedure, where we first train the energy-based model to perform implicit density estimation and subsequently train the whole model by maximizing the objective in Eq.~(\ref{eq:gedi_obj}).

Regarding the first stage, we maximize only the generative term in Eq.~(\ref{eq:gedi_obj}) whose gradient is given by the following relation:
\begin{align}
    -\nabla_\Psi CE(p,p_\Psi) &= \sum_{i=1}^n\mathbb{E}_{p(x_i)}\left\{\nabla_\Psi\log e^{u^Tenc(x_i)}\right\} \nonumber\\
    &\quad-n\nabla_\Psi\log\Gamma(\Psi) \nonumber\\
    &=\sum_{i=1}^n\mathbb{E}_{p(x_i)}\{\nabla_\Psi\log e^{u^Tenc(x_i)}\} \nonumber\\
    &\quad n\mathbb{E}_{p_\Psi(x)}\left\{\nabla_\Psi\log e^{u^Tenc(x)}\right\}
    \label{eq:energy_obj}
\end{align}
where the first and the second expectations in Eq.~(\ref{eq:energy_obj}) are estimated using the training and the generated data, respectively. To generate data from $p_\Psi$, we use a sampler based on Stochastic Gradient Langevin Dynamics (SGLD), thus following recent best practices to train energy-based models~\cite{xie2016theory,nijkamp2019learning,du2019implicit,nijkamp2020anatomy}.

\begin{algorithm}[t]
 \caption{GEDI Training.}
 \label{alg:gedi}
    \textbf{Input:} $x_{1:n}$, $\text{Iters}_1$, $\text{Iters}_2$, $T$, $\epsilon$, SGLD and Adam optimizer hyperparameters\;
    \textbf{Output:} Trained model $g$\;
    \# Step 1\;
    \textbf{For} $\text{iter}=1,\dots,\text{Iters}_1$\;
    \qquad Generate samples from $p_\Psi$ using SGLD\;
    \qquad $\Psi\leftarrow\text{Adam}$ maximizing $-CE(p,p_\Psi)$\;
    \# Step 2\;
    \textbf{For} $\text{iter}=1,\dots,\text{Iters}_2$\;
    \qquad $x_{1:n}''\leftarrow\text{DAM}(x_{1:n},p_\Psi,\epsilon,T)$\;
    \qquad Generate samples from $p_\Psi$ using SGLD\;
    \qquad $\Psi,\Theta\leftarrow\text{Adam}$ maximizing Eq.~(\ref{eq:gedi_obj})\;
    \textbf{Return} $g$\;
\end{algorithm}
Regarding the second stage, we maximize the whole objective in Eq.~(\ref{eq:gedi_obj}). Specifically, at each training iteration, we run the DAM routine to obtain the augmented samples $x_{1:n}''$ and compute the last addend for the discriminative part in Eq.~(\ref{eq:gedi_obj}), viz. $\mathcal{L}_{DI}(\Theta;p_\Psi)$. For both $\mathcal{L}_{DI}(\Theta;\mathcal{T})$ and $\mathcal{L}_{DI}(\Theta;p_\Psi)$ we use the same differentiable clustering strategy used in SwAV~\cite{caron2020unsupervised}. The whole learning procedure is summarized in Algorithm~\ref{alg:gedi}.

\textbf{Computational requirements.} Compared to traditional SSL training, and more specifically to SwAV, our learning algorithm requires additional operations and therefore increased computational requirements (but constant given $T$ and $\text{Iters}_1$). Indeed, (i) we need an additional training step (Step 1) to pre-train a generative model to approximate the unknown data density and to ensure the proper working of DAM in the second step, (ii) we need to generate samples from $p_\Psi$ to continue learning the generative model in Step 2 and (ii) we also need to run $T$ additional forward and backward passes through the energy-based model to run the DAM strategy at each iteration of the GEDI training.

\section{Related Work}
We organize the related work according to different objective categories, namely contrastive, cluster-based and non-contrastive self-supervised approaches. Additionally, we discuss recent theoretical results, augmentation strategies as well as connections to energy-based models. For an exhaustive overview of self-supervised learning, we invite the interested reader to check two recent surveys~\cite{jing2021self,liu2021self}.

\textbf{Contrastive objectives and connection to mutual information.} Contrastive learning represents an important family of self-supervised learning algorithms, which is rooted in the maximization of the mutual information between the data and its latent representation~\cite{linsker1988self,becker1992self}. Estimating and optimizing mutual information from samples is notoriously difficult~\cite{mcallester2020formal}, especially when dealing with high-dimensional data. Most recent approaches focus on devising variational lower bounds on mutual information~\cite{agakov2004algorithm}. Indeed, several popular objectives, like the mutual information neural estimation (MINE)~\cite{belghazi2018mutual}, deep InfoMAX~\cite{hjelm2018learning}, noise contrastive estimation (InfoNCE)~\cite{oord2018representation,henaff2020data,chen2020simple,lee2022prototypical,xu2022k} to name a few,  all belong to the family of variational lower bounds~\cite{poole2019variational}. All these estimators have different properties in terms of bias-variance trade-off~\cite{tschannen2019mutual,song2020understanding}. Our work model contrastive learning using an equivalent probabilistic graphical model and a corresponding objective 
 function based on the data log-likelihood, thus providing an alternative view to the principle of mutual information maximization. This is similar in spirit to the formulations proposed in the recent works of~\cite{mitrovic2021representation,tomasev2022pushing,scherr2022self,xu2022k}. Differently from these works, we are able to extend the log-likelihood perspective to other families of self-supervised approaches and also to highlight and exploit their connections to energy-based models.

\textbf{Discriminative/Cluster-based objectives}. There are also recent advances in using clustering techniques in representation learning. For example, DeepCluster~\cite{caron2018deep} uses k-means and the produced cluster assignments as pseudo-labels to learn the representation. The work in~\cite{huang2022learning} introduces an additional regularizer for deep clustering, invariant to local perturbations applied to the augmented latent representations. The work in~\cite{asano2019self} shows that the pseudo-label assignment can be seen as an instance of the optimal transport problem. SwAV~\cite{caron2020unsupervised} proposes to use the Sinkhorn-Knopp algorithm to optimize the optimal transport objective~\cite{cuturi2013sinkhorn} and to perform a soft cluster assignment. Finally, contrastive clustering~\cite{li2021contrastive} proposes to minimize the optimal transport objective in a contrastive setting, leveraging both positive and negative samples. Our work provides a simple yet concise formulation of cluster-based self-supervised learning based on the principle of likelihood maximization. Additionally, thanks to the connection with energy-based models, we can perform implicit density estimation and leverage the learnt information to improve the clustering performance.

\textbf{Non-contrastive objectives.}
It's important to mention that new self-supervised objectives have been emerged recently~\cite{zbontar2021barlow,grill2020bootstrap} as a way to avoid using negative samples, which are typically required by variational bouns on the mutual information, namely BYOL~\cite{grill2020bootstrap}, SimSiam~\cite{chen2020simple}, DINO~\cite{caron2021emerging}, Zero-CL~\cite{zhang2021zero}, W-MSE~\cite{ermolov2021whitening}, Barlow Twins~\cite{zbontar2021barlow}, VICReg~\cite{bardes2022vicreg} and its variants~\cite{bardes2022vicregl}, MEC~\cite{liu2022self} and CorInfoMax~\cite{ozsoy2022self}.
DINO proposes to maximize a cross-entropy objective to match the probabilistic predictions from two augmented versions of the same image. BYOL, SimSiam, W-MSE consider the cosine similarity between the embeddings obtained from the augmented pair of images. Additionally, W-MSE introduces a hard constraint implemented as a differentiable layer to whiten the embeddings. Similarly, Barlow Twins proposes a soft whitening by minimizing the Frobenius norm between the cross-correlation matrix of the embeddings and the identity matrix. Zero-CL pushes forward the idea of whitening the features by also including an instance decorrelation term. VICReg and its variant extend over Barlow Twins by computing the sample covariance matrix, instead of the correlation one (thus avoiding to use batch normalization), by enforcing an identity covariance and by introducing an additional regularizer term to minimize the mean squared error between the embeddings of the two networks and to promote the invariance of the embeddings. Similarly, the work in~\cite{tomasev2022pushing} uses an invariance loss function in conjunction to the contrastive InfoNCE objective. MEC maximizes the log-determinant of the covariance matrix for the latent representation, thus promoting maximum entropy under the Gaussian assumption. Additionally, CorInfoMax extends MEC by introducing a term enforcing the representation to be invariant under data augmentation. In Section 2.3, we can cast the non-contrastive problem as a minimization of the Kullback Leibler divergence between the latent posterior and a standard normal density prior. In essence, our work allows to compactly represent the family of non-contrastive methods using a likelihood-based objective.

\textbf{Additional objectives.} Several works have investigated the relation between different families of self-supervised approaches, thus leading to hybrid objective functions~\cite{von2021self,garrido2022duality}. In contrast, our work attempts to provide a unified view from a probabilistic perspective and to highlight/exploit its connection to energy-based models.

The main idea of generative and discriminative training originally appeared in the context of Bayesian mixture models~\cite{sansone2016classtering}. Specifically, the authors proposed to jointly learn a generative model and cluster data in each class in order to be able to discover subgroups in breast cancer data. Subsequently, the work in~\cite{liu2020hybrid} pushed the idea forward and apply it to a supervised deep learning setting. Instead, our work focuses on self-supervised learning and on generative models, thus avoiding the need for additional supervision on the categorical variable $y$. Recently, the work in~\cite{li2022neural} uses the maximum coding rate criterion to jointly learn an embedding and cluster it. However, the training proceeds in a multi-stage fashion. In contrast, our work provides a simple formulation enabling to jointly learn and cluster the embeddings in one shot. Additionally, we can leverage the information from a learnt energy-based model to further boost the self-supervised learning performance.

\textbf{Theory of self-supervised learning.} 
Several works have theoretically analysed self-supervised approaches, both for contrastive~\cite{saunshi2019theoretical,wang2020understanding,tosh2021contrastive,haochen2021provable,saunshi2022understanding} and non-contrastive methods~\cite{tian2021understanding,kang2022bridging,weng2022investigation,wen2022mechanism}, to motivate the reasons for their successes, identify the main underlying principles and subsequently provide more principled/simplified solutions. Regarding the former family of approaches, researchers have (i) identified key properties, such as representation alignment (i.e. feature for positive pairs need to be close to each other) and uniformity (to avoid both representational and dimensional collapse)~\cite{wang2020understanding}, (ii) formulated and analyzed the problem using data augmentation graphs~\cite{haochen2021provable} and (iii) looked into generalization bounds on the downstream supervised performance~\cite{saunshi2022understanding,bao2022surrogate}. Regarding the latter family of approaches, the main focus has been devoted to understand the reasons on why non-contrastive approaches avoid trivial solutions. In this regard, asymmetries, in the form of stop-gradient and diversified predictors, are sufficient to ensure well-behaved training dynamics~\cite{tian2021understanding,weng2022investigation,wen2022mechanism}. Importantly, the asymmetries are shown to implicitly constrain the optimization during training towards solutions with decorrelated features~\cite{kang2022bridging}.  Recent works have also looked at identifying connections between contrastive and non-contrastive methods~\cite{dubois2022improving,garrido2022duality,balestriero2022contrastive} to unify the two families. The work in~\cite{dubois2022improving} proposes a set of desiderata for representation learning, including large dimensional representations, invariance to data augmentations and the use of at least one linear predictor to ensure good performance on linear probe evaluation tasks. The work in~\cite{garrido2022duality} analyzes the relations between contrastive and non-contrastive methods, showing their similarities and differences. Both family of approaches learn in a contrastive manner. However, while contrastive solutions learn by contrasting between samples, non-contrastive ones focus on contrasting between the dimensions of the embeddings. The work in~\cite{balestriero2022contrastive} studies the minima in terms of representations for the different loss functions proposed in the two families. The authors are able to show that such minima are equivalent to solutions achieved by spectral methods. This provides additional evidence on the similarities between the approaches and the possibility for their integration.

Our work provides a unifying view of the different classes, allowing to derive several loss functions in a principled manner using variational inference on the data log-likelihood. Additionally, while previous approaches intrinsically rely on a crude approximation/discretization of the data manifold, namely the data augmentation graph~\cite{balestriero2022contrastive}, our framework allows to leverage the generative perspective to improve the approximation of the manifold structure and consequently learn a better symbolic representation (where symbols refer to clusters).

\textbf{Augmentation strategies for self-supervised learning.} Augmentations play a crucial role for representation learning and have been extensively studied. Here, we focus only on most recent and related approaches. The work in~\cite{wang2022contrastive} distinguishes between weak and strong augmentations (i.e the former ones are based on cropping, resize, color jitter etc, while the latter ones are based on shear, rotation, equalization and so on), thus showing the benefits of exploiting both of them. The works in~\cite{yang2022identity,shi2022adversarial} frame the problem of data augmentation in an adversarial setting. The first approach~\cite{yang2022identity} performs augmentation in the latent space by looking for small perturbations leading to large changes in the input space. The second approach~\cite{shi2022adversarial} learns to mask the data in an adversarial setting, thus enabling to learn more semantically meaningful features. Finally, the work in~\cite{li2022metaug} proposes to learn a generator directly in the latent space using a meta-learning procedure. This allows to produce more informative augmentations and consequently reducing the batch size typically required by contrastive approaches. Our work complements all these previous solutions with an augmentation strategy, that takes into account the information about the underlying manifold structure, thus leveraging connectedness properties.

\textbf{Energy-based models.} 
Recently, works have considered the use of self-supervised learning~\cite{lecun2022path} for the purposes of Out-of-Distribution detection~\cite{hendrycks2019using,winkens2020contrastive,mohseni2020self}. This is a common characteristics of energy-based models and indeed our work highlight the connection of self-supervised learning to these models. To the best of our knowledge, there is only one recent work exploring the integration between self-supervised learning approaches and energy models~\cite{kim2022energy}. The authors propose to use an energy-based model to learn a joint distribution over the two augmented views for the same data. The resulting objective can be decomposed into a conditional distribution term, leading to a contrastive learning criterion, and a marginal distribution term, leading to an energy-based model criterion. Therefore, the work only considers the integration between contrastive methods and energy-based models. In contrast, our work goes a step forward by showing a general methodology to integrate generative and SSL approaches and consequently proposing an instantiation of the GEDI framework, based on energy-based and cluster-based SSL models.

\section{Experiments}
We perform experiments to evaluate the discriminative performance of GEDI and its competitors, namely an energy-based model JEM~\cite{grathwohl2020your}, which is trained with persistent contrastive divergence (similarly to our approach) and 2 self-supervised baselines, viz. a negative-free approach based on Barlow Twins~\cite{zbontar2021barlow} and a discriminative one based on SwAV~\cite{caron2020unsupervised}.
The whole analysis is divided into three main experimental settings, the first one based on two synthetic datasets, including moons and circles, the second one based on real-world data, including SVHN, CIFAR-10 and CIFAR-100, and the last one based on a neural-symbolic learning task in the small data regime constructed from MNIST. We use existing code both as a basis to build our solution and also to run the experiments for the different baselines. In particular, we use the code from~\cite{duvenaud2021no} for training energy-based models and the repository from~\cite{costa2022solo} for all self-supervised baselines. Our code will be publicly released in its entirety upon acceptance. Implementation details as well as additional experiments are reported in the Supplementary Material.

\subsection{Synthetic data}
\begin{table}[t]
  \caption{Clustering performance in terms of normalized mutual information (NMI) on test set (moons and circles). Higher values indicate better clustering performance. Mean and standard deviations are computed from 5 different runs. GEDI uses $T=10$ moves in DAM.}
  \label{tab:nmi_toy}
  \centering
\begin{tabular}{@{}lrrrr@{}}
\toprule
\textbf{Dataset} & \textbf{JEM} & \textbf{SwAV} & \textbf{GEDI} & \textbf{Gain}  \\
\midrule
Moons & 0.0$\pm$0.0 & 0.8$\pm$0.2 & \textbf{0.98}$\pm$\textbf{0.02} & \textbf{+0.18} \\
Circles & 0.0$\pm$0.0 & 0.0$\pm$0.0 & \textbf{1.00}$\pm$\textbf{0.01} & \textbf{+1.00} \\
\bottomrule
\end{tabular}
\end{table}
We consider two well-known synthetic datasets, namely moons and circles. We use a multi-layer perceptron (MLP) with two hidden layers (100 neurons each) for $enc$ and one with a single hidden layer (4 neurons) for $proj$, we choose $h=2$ and we choose $\mathcal{T}(x'|x)=N(0, \sigma^2 I)$ with $\sigma=0.03$ as our data augmentation strategy. We train JEM, SwAV and GEDI for $7k$ iterations using Adam optimizer with learning rate $1e-3$. Further details about the hyperparameters are available in the Supplementary Material (Section G). We evaluate the clustering performance both qualitatively, by visualizing the cluster assignments using different colors, as well as quantitatively, by using the Normalized Mutual Information (NMI) score. Furthermore, we conduct an ablation study for the different components of GEDI.

\begin{figure}
     \centering
     \begin{subfigure}[b]{0.3\linewidth}
         \centering
         \includegraphics[width=0.9\textwidth]{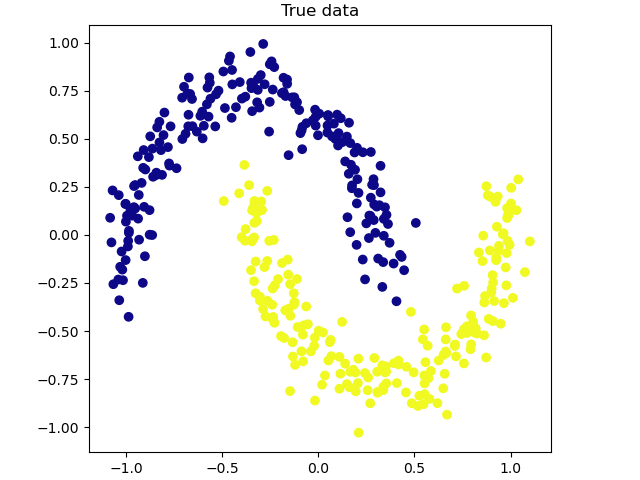}
         \caption{Ground truth}
     \end{subfigure}%
     \begin{subfigure}[b]{0.3\linewidth}
         \centering
         \includegraphics[width=0.9\textwidth]{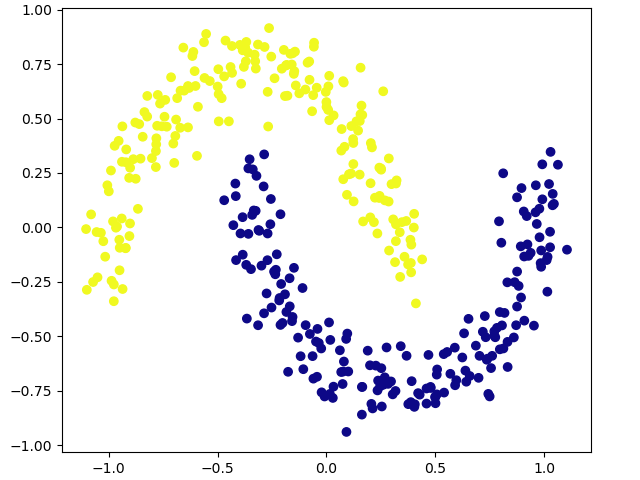}
         \caption{SwAV}
     \end{subfigure}%
     \begin{subfigure}[b]{0.3\linewidth}
         \centering
         \includegraphics[width=0.9\textwidth]{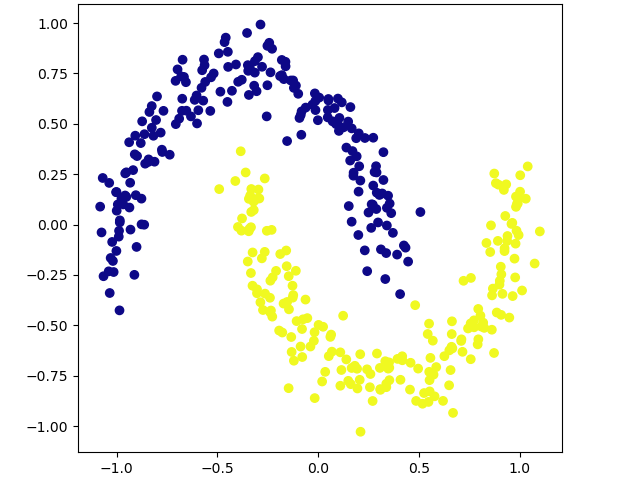}
         \caption{GEDI}
     \end{subfigure}%
     \\
    \begin{subfigure}[b]{0.3\linewidth}
         \centering
         \includegraphics[width=0.9\textwidth]{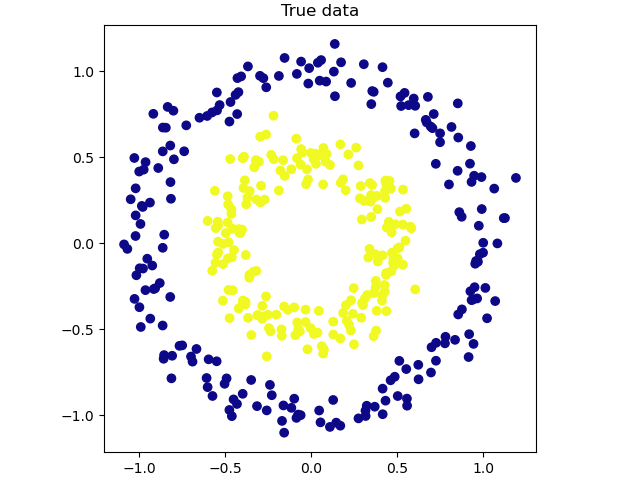}
         \caption{Ground truth}
     \end{subfigure}%
     \begin{subfigure}[b]{0.3\linewidth}
         \centering
         \includegraphics[width=0.9\textwidth]{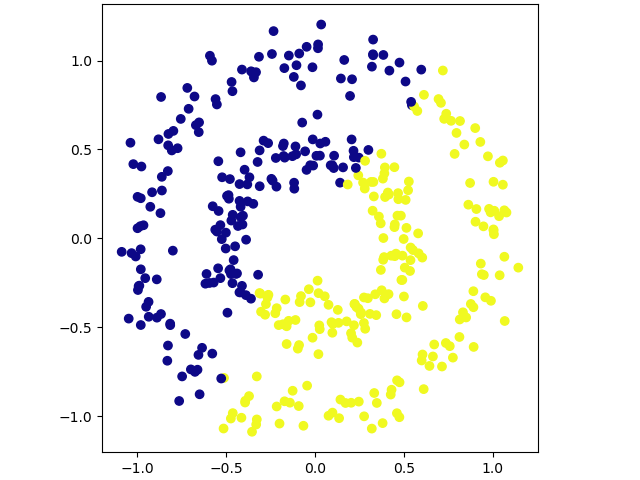}
         \caption{SwAV}
     \end{subfigure}%
     \begin{subfigure}[b]{0.3\linewidth}
         \centering
         \includegraphics[width=0.9\textwidth]{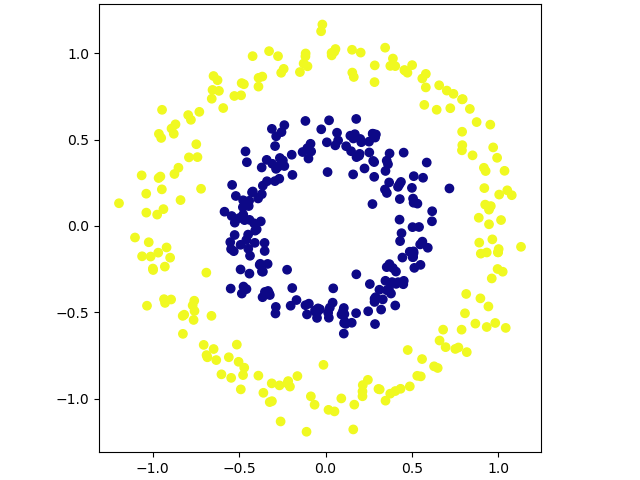}
         \caption{GEDI}
     \end{subfigure}%
     \caption{Qualitative visualization of the clustering performance for the different strategies (only SwAV and GEDI are shown) on moons (a-c) and on circles (d-f) datasets. Colors identify different cluster predictions.}
     \label{fig:labels_toy}
\end{figure}

We report all quantitative performance in Table~\ref{tab:nmi_toy}. Specifically, we observe that JEM fails to solve the clustering task for both datasets. This is quite natural, as JEM is a purely generative approach, mainly designed to perform implicit density estimation. SwAV can only solve the clustering task for the moons dataset, highlighting the fact that it is not able to exploit the information from the underlying density used to generate the data. However, GEDI can recover the true clusters in both datasets. This is due to the fact that GEDI uses the information from the generative component through DAM to inform the cluster-based one. Consequently, GEDI is able to exploit the manifold structure underlying data. Figure~\ref{fig:labels_toy} provide some examples of predictions by SwAV and GEDI on the two datasets. From the figure, we can clearly see that GEDI can recover the two data manifolds up to a permutation of the labels.

\begin{figure}
     \centering
     \begin{subfigure}[b]{0.49\linewidth}
         \centering
         \includegraphics[width=0.9\textwidth]{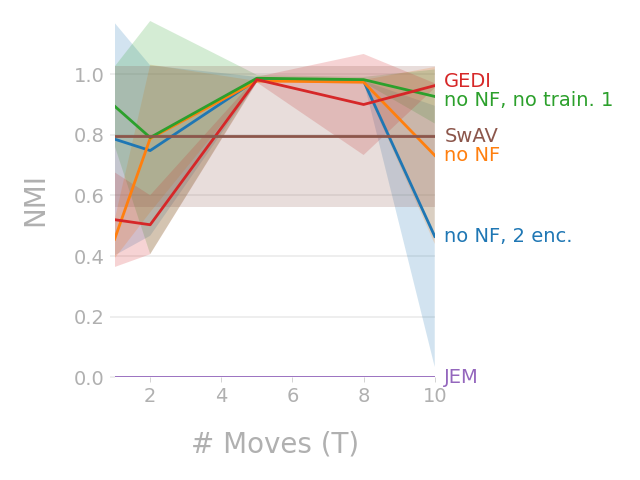}
         \caption{Moons}
     \end{subfigure}%
     \begin{subfigure}[b]{0.49\linewidth}
         \centering
         \includegraphics[width=0.9\textwidth]{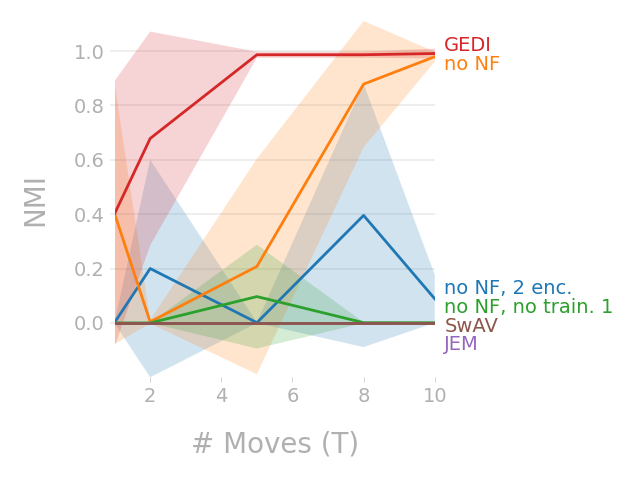}
         \caption{Circles}
     \end{subfigure}%
     \caption{Ablation study for GEDI on moons and circles dataset. The curves are obtained by choosing different values of $T$ for DAM, namely $T\in\{1,2,5,8,10\}$.}
     \label{fig:ablation_toy}
\end{figure}

We conduct an ablation study to understand the impact of the different components of GEDI. We compare four different versions of GEDI, namely the full version (called simply \textit{GEDI}), GEDI trained without $\mathcal{L}_{NF}(\Theta)$ (called \textit{no NF}), GEDI trained without the first stage and also without $\mathcal{L}_{NF}(\Theta)$ (called \textit{no NF, no train. 1}) and GEDI trained without $\mathcal{L}_{NF}(\Theta)$ using two different encoders for computing the discriminative and the generative terms in our objective (called \textit{no NF, 2 enc.}). From the results in Figure~\ref{fig:ablation_toy}, we can make the following observations: (i) DAM plays a crucial role to recover the manifold structure, as the performance increase with the number of steps $T$ for \textit{GEDI}, \textit{no NF} and \textit{no NF, 2 enc.}. However, the strategy is not effective when we don't use the first training stage, as demonstrated by the results obtained by \textit{no NF, no train. 1} on circles. This confirms our original hypothesis that DAM requires a good generative model; (ii) When looking at the results obtained by \textit{no NF} and \textit{no NF, 2 enc.}, we observe that there is a clear advantage, especially on circles, by using a single encoder, highlighting the fact that the integration between generative and self-supervised models is effective not only at the objective level but also at the architectural one; (iii) From the comparison between \textit{GEDI} and \textit{no NF} on circles, we observe improved performance and reduced variance. This suggests that $\mathcal{L}_{NF}(\Theta)$ complements the other terms in our objective and therefore contributes to driving the learning towards desired solutions. Additional analysis is provided in Section H of the Supplementary Material.

\begin{figure}
     \centering
     \begin{subfigure}[b]{0.45\linewidth}
         \centering
         \includegraphics[width=0.9\textwidth]{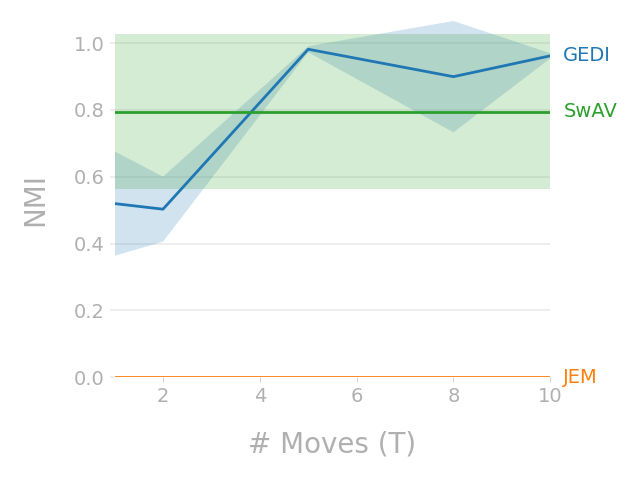}
         \caption{Moons}
     \end{subfigure}%
     \begin{subfigure}[b]{0.45\linewidth}
         \centering
         \includegraphics[width=0.9\textwidth]{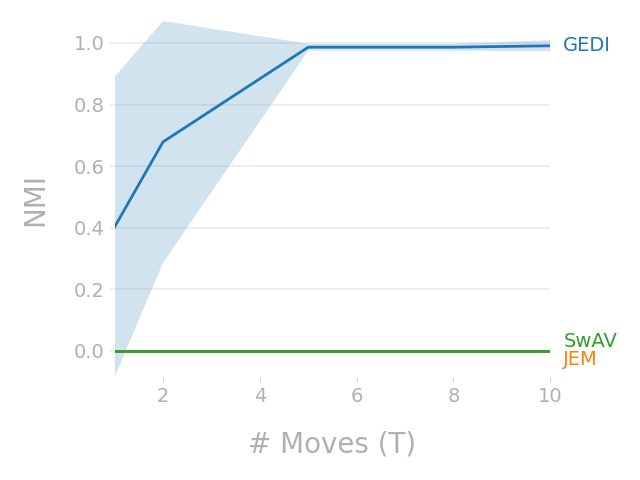}
         \caption{Circles}
     \end{subfigure}%
     \caption{NMI achieved by JEM, SwAV and GEDI on moons and circles dataset. The curves are obtained by choosing different values of $T$ for DAM, namely $T\in\{1,2,5,8,10\}$.}
     \label{fig:nmi_toy}
\end{figure}

\subsection{SVHN, CIFAR-10, CIFAR-100}\label{sec:experiments_real}
\begin{table}[t]
  \caption{Clustering performance in terms of normalized mutual information on test set (SVHN, CIFAR-10, CIFAR-100). Higher values indicate better clustering performance. We observe unstable training for SwAV on CIFAR-100. We report the best performance achieved out of 10 experiments.}
  \label{tab:nmi_real}
  \centering
\begin{tabular}{@{}lrrrrr@{}}
\toprule
\textbf{Dataset} & \textbf{JEM} & \textbf{Barlow} & \textbf{SwAV} & \textbf{GEDI} & \textbf{Gain}  \\
\midrule
SVHN & 0.04 & 0.20 & 0.24 & \textbf{0.39} & \textbf{+0.15} \\
CIFAR-10 & 0.04 & 0.22 & 0.39 & \textbf{0.41} & \textbf{+0.02}\\
CIFAR-100 & 0.05 & 0.46 & 0.69$^*$ & \textbf{0.72} & \textbf{+0.03} \\
\bottomrule
\end{tabular}
\end{table}
\begin{table}[t]
  \caption{Ablation study for clustering performance in terms of normalized mutual information on test set (SVHN, CIFAR-10, CIFAR-100). Higher values indicate better clustering performance.}
  \label{tab:ablation_real}
  \centering
\begin{tabular}{@{}lrrrr@{}}
\toprule
\textbf{Dataset} & \textbf{no NF, 2 enc.} & \textbf{no NF} & \textbf{GEDI} & \textbf{Gain}  \\
\midrule
SVHN & 0.21 & 0.31 & \textbf{0.39} & \textbf{+0.08} \\
CIFAR-10 & 0.38 & \textbf{0.40} & \textbf{0.41} & +0.00\\
CIFAR-100 & \textbf{0.75} & \textbf{0.76} & 0.72s & -0.04 \\
\bottomrule
\end{tabular}
\end{table}
We consider three well-known computer vision benchmarks, namely SVHN, CIFAR-10 and CIFAR-100. We use a simple 8-layer Resnet network for the backbone encoder for both SVHN and CIFAR-10 (around 1M parameters) and increase the hidden layer size for CIFAR-100 (around 4.1M parameters) as from~\cite{duvenaud2021no}. We use a MLP with a single hidden layer for $proj$ (the number of hidden neurons is double the number of inputs), we choose $h=256$ for CIFAR-100 and $h=128$ for all other cases. Additionally, we use data augmentation strategies commonly used in the SSL literature, including color jitter, and gray scale conversion to name a few. We train JEM, Barlow, SwAV and GEDI for $100$ epochs using Adam optimizer with learning rate $1e-4$ and batch size $64$. Further details about the hyperparameters are available in the Supplementary Material (Section H). Similarly to the toy experiments, we evaluate the clustering performance by using the Normalized Mutual Information (NMI) score and also add the ablation experiments for GEDI.

From Table~\ref{tab:nmi_real}, we observe that GEDI is able to outperform all other competitors by a large margin, thanks to the properties of both generative and self-supervised models. This provides real-world evidence on the benefits of the proposed unification. The ablation study in Table~\ref{tab:ablation_real} confirms the findings observed for the toy experiments. In the Supplementary Material (Section J), we provide additional evaluation on linear probe evaluation, generation and OOD detection tasks.

\subsection{Neural-symbolic setting}
For the final task, we consider applying the self-supervised learning approach to a neural-symbolic setting. For this, we borrow an experiment from DeepProbLog \cite{manhaeve2018deepproblog}. In this task, each example consists of a three MNIST images such that the value of the last one is the sum of the first two, e.g. $\digit{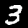} + \digit{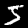} = \digit{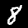}$. This can thus be considered a minimal neural-symbolic tasks, as it requires a minimal reasoning task (a single addition) on top of the image classification task. This task only contains positive examples. 
We use the inference mechanism from DeepProbLog to calculate the probability that this sum holds, and optimize this probability using the cross-entropy loss function, which is optimized along with the other loss functions. For this setting, this coincides with the Semantic Loss function~\cite{xu2018semantic}. To be able to calculate the probability of this addition constraint, we need the classification probabilities for each digit.

It is a specifically interesting use case for the self-supervised learning, since when only the probability is optimized, the neural network tends to collapse onto the trivial solution of classifying each digit as a $0$, as shown in \cite{manhaeve2018deepproblog}. This is a logically correct solution, but an undesirable solution. Optimizing the discriminative objective should prevent this collapse. We hypothesize that a neural network can be trained to correctly classify MNIST digits by using the SSL training objective and the logical constraint. Since the MNIST dataset is an easy dataset, we focus on the small data regime, and see whether the logical constraint is able to provide additional information.
The hyperparameters are identical to those used in Section~\ref{sec:experiments_real}. Further details about the hyperparameters are dependent on the data regime, and are available in Appendix~K. 

We evaluate the model by measuring the accuracy and NMI of the ResNet model on the MNIST test dataset for different numbers of training examples. The results are shown in Table~\ref{tab:nesy_result}.  Here, N indicates the number of addition examples, which each have $3$ MNIST digits.
As expected, the DeepProbLog baseline from \cite{manhaeve2018deepproblog} completely fails to classify MNIST images. It has learned to map all images to the class $0$, as this results in a very low loss when considering only the logic, resultsing in an accuracy of $0.10$ and an NMI of $0.0$.
The results also show that, without the NeSy constraint, the accuracy is low for all settings. The NMI is higher, however, and increases as there is more data available. This is expected, since the model is still able to learn how to cluster from the data. However, it is  unable to correctly classify, as there is no signal in the data that is able to assign the correct label to each cluster. By including the constraint loss, the accuracy improves, as the model now has information on which cluster belongs to which class. Furthermore, it also has a positive effect on the NMI, as we have additional information on the clustering which is used by the model.

These results show us that the proposed method is beneficial to learn to correctly recognize MNIST images using only a weakly-supervised constraint, whereas other NeSy methods fail without additional regularization. Furthermore, we show that the proposed method can leverage the information offered by the constraint to further improve the NMI and classification accuracy.

\begin{table*}[ht]
\caption{The accuracy and NMI of GEDI on the MNIST test set after training on the addition dataset, both with and without the NeSy constraint. Additionally, we use DeepProbLog~\cite{manhaeve2018deepproblog} as a baseline without using our GEDI model. We trained each model 5 times and report the mean and standard deviation.}
\label{tab:nesy_result}
\centering
\begin{tabular}{@{}rrrrrrr@{}}
 \toprule
 & \multicolumn{2}{c}{\textbf{Without GEDI}} & \multicolumn{2}{c}{\textbf{Without constraint}} &  \multicolumn{2}{c}{\textbf{With constraint}} \\
\textbf{N}  & 
\textbf{Acc.}               & \textbf{NMI}                &
\textbf{Acc.}               & \textbf{NMI}               &  \textbf{Acc.}             & \textbf{NMI}              \\ 
\midrule
100
& $0.10 \pm 0.00$ & $0.00 \pm 0.00$
& $0.08 \pm 0.03$ & $0.28 \pm 0.03$ & $\mathbf{0.25 \pm 0.03}$ & $\mathbf{0.41 \pm 0.03}$ \\
1000
& $0.10 \pm 0.00$ & $0.00 \pm 0.00$
& $0.09 \pm 0.02$ & $0.47 \pm 0.10$ & $\mathbf{0.52 \pm 0.26}$ & $\mathbf{0.86 \pm 0.06}$ \\
10000
& $0.10 \pm 0.00$ & $0.00 \pm 0.00$
& $0.17 \pm 0.12$ & $0.68 \pm 0.09$ & $\mathbf{0.98 \pm 0.00}$ & $\mathbf{0.97 \pm 0.01}$ \\
\bottomrule
\end{tabular}
\end{table*}

\section{Conclusions and Future Research}
In this study, we have presented a unified framework for combining generative and self-supervised learning models, and demonstrated its effectiveness through an implementation that outperforms existing cluster-based self-supervised learning strategies by a wide margin. However, this solution comes at the cost of increased computational requirements. Specifically, it requires an additional training phase to learn an energy-based model and additional forward and backward passes through the generative component in order to use the DAM procedure. Future work will focus on developing strategies that require only a single training phase and reduce the number of evaluations of the generative model.

Since our work bridges the fields of self-supervised learning and generative models, we expect to see follow-up studies proposing new implementations of the GEDI framework, such as new ways of integrating self-supervised learning and latent variable models. Additionally, we have shown the benefits of considering the manifold structure of data through the DAM procedure, and believe that further progress can be made by combining self-supervised learning with other areas of mathematics such as topology and differential geometry. Finally, we have demonstrated that the GEDI solution can be easily integrated into existing statistical relational reasoning frameworks, paving the way for new neuro-symbolic integrations and enabling the handling of low data regimes, which are currently beyond the reach of existing self-supervised learning solutions.





%

\ifCLASSOPTIONcompsoc
  \section*{Acknowledgments}
\else
  \section*{Acknowledgment}
\fi
The authors would like to thank Michael Puthawala for insightful discussion on GEDI and DAM.

This research is funded by TAILOR, a project from the EU Horizon 2020 research and innovation programme under GA No 952215. This research received also funding from the Flemish Government under the “Onderzoeksprogramma Artificiële Intelligentie (AI) Vlaanderen” programme.

ES proposed the theory of GEDI and DAM, implemented the solution for the toy and real cases and wrote the paper. RM implemented and wrote the experiments for the neuro-symbolic setting and also helped by proof-reading the article.

\ifCLASSOPTIONcaptionsoff
  \newpage
\fi

\bibliographystyle{IEEEtran}
\bibliography{ref}

%

\begin{IEEEbiography}[{\includegraphics[width=1in,height=1.25in,clip,keepaspectratio]{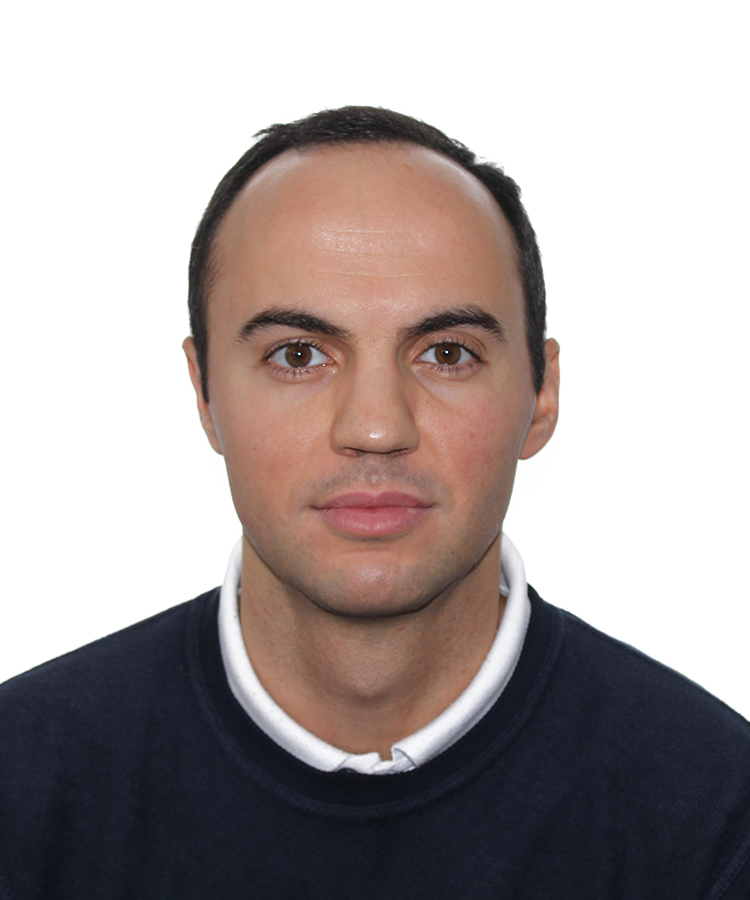}}]{Emanuele Sansone} obtained his PhD degree in 2018 from the University of Trento, where he wrote his thesis titled "Towards Uncovering the True Use of Unlabeled Data in Machine Learning". During his PhD, he visited Nanjing University in China and worked on positive unlabeled learning under the supervision of Prof. Zhi-Hua Zhou. After completing his PhD, Sansone moved to London and became a permanent research scientist at Huawei, where he worked on topics related to generative models and representation learning. In 2020, he returned to academia and joined KU Leuven in Belgium as a postdoctoral researcher, working in the group of Prof. Luc De Raedt. Sansone's research interests lie at the intersection of unsupervised and statistical relational learning. He has served as a reviewer for top-tier journals and conferences in the field of machine learning and artificial intelligence. He has received several awards for his professional service, including the outstanding reviewer award at ICLR 2021, ICML 2022, NeurIPS 2022, AISTATS 2023.
\end{IEEEbiography}

\begin{IEEEbiography}[{\includegraphics[width=1in,height=1.25in,clip,keepaspectratio]{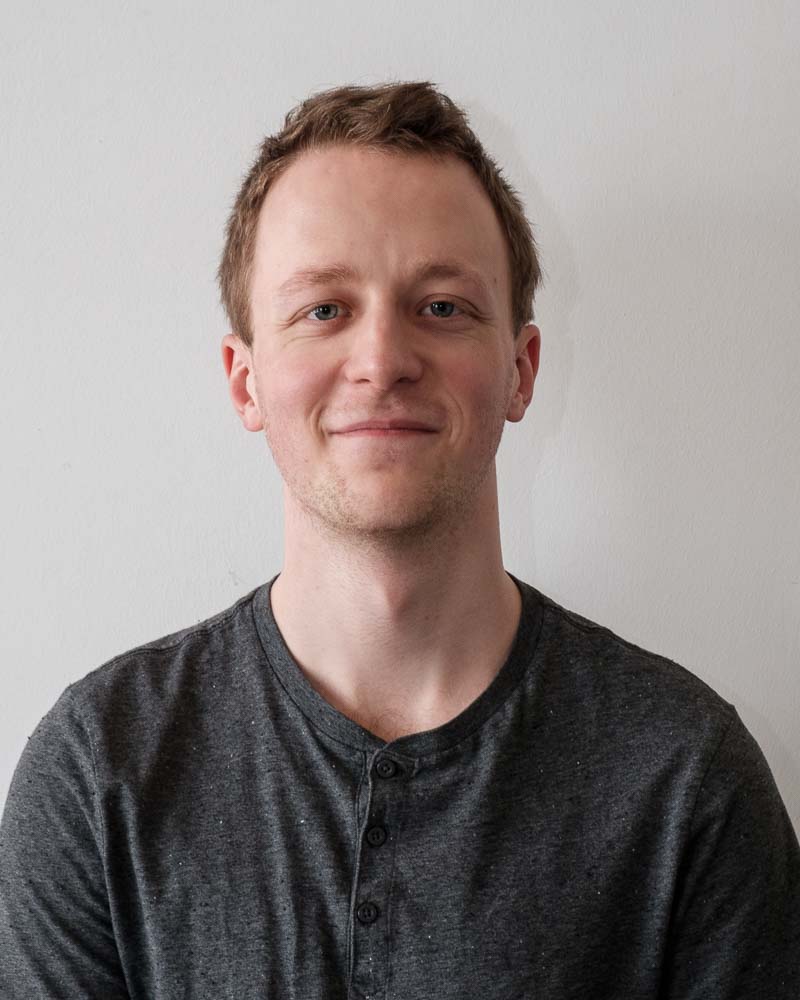}}]{Robin Manhaeve}
obtained his PhD degree in 2021 at the KU Leuven with an SB Fellowship from the Research Foundation – Flanders (FWO), under the supervision of Prof. Luc De Raedt in the DTAI research group. His PhD thesis is titled "Neural Probabilistic Logic Programming. After graduating, he continued working under the supervision of Prof. De Raedt as a postdoctoral researcher.
His main resaerch interests lie in the field of neural-symbolic integration, more specifically in the integration of neural networks and probabilistic logic programming.
\end{IEEEbiography}

\appendices
{\section{Alternative View of Contrastive SSL}\label{sec:A}}
Let us focus the analysis on a different graphical model from the one in Section 2.1, involving an input vector $x$ and a latent embedding $z$. Without loss of generality, we can discard the index $i$ and focus on a single observation. We will later extend the analysis to the multi-sample case. Now, consider the conditional distribution of $x$ given $z$ expressed in the form of an energy-based model $p(x|z;\Theta)=\frac{e^{f(x,z)}}{\Gamma(z;\Theta)}$, where $f:\Omega\times\mathcal{S}^{h-1}\rightarrow\mathbb{R}$ is a score function and $\Gamma(z;\Theta)=\int_\Omega e^{f(x,z)}dx$ is the normalizing factor.\footnote{We assume that $f$ is a well-behaved function, such that the integral value $\Gamma(z;\Theta)$ is finite for all $z\in\mathcal{S}^{h-1}$.} Based on this definition, we can obtain the following  lower-bound on the data log-likelihood:
\begin{align}
    \mathbb{E}_{p(x)}\{\log p(x)\} &= KL(p(x)\|p(x;\Theta)) + \mathbb{E}_{p(x)}\{\log p(x;\Theta)\} \nonumber\\
    &\geq \mathbb{E}_{p(x)}\{\log p(x;\Theta)\} \nonumber \\
    &\geq \mathbb{E}_{p(x)q(z|x)}\{\log p(x|z;\Theta)\}\nonumber\\
    &\qquad-\mathbb{E}_{p(x)}\{KL(q(z|x)\|p(z;\Theta))\}  \nonumber\\
    &= \mathbb{E}_{p(x)q(z|x)}\{\log p(x|z;\Theta)\} \nonumber\\
    &\qquad+ \mathbb{E}_{p(x)q(z|x)}\{\log p(z;\Theta)\} \nonumber\\
    &= \mathbb{E}_{p(x)q(z|x)}\{f(x,z)\} \nonumber\\
    &\qquad+ \mathbb{E}_{p(x)q(z|x)}\bigg\{\log\frac{p(z;\Theta)}{\Gamma(z;\Theta)}\bigg\}\nonumber\\
    &\doteq \text{ELBO}_{EBM}
    \label{eq:generative}
\end{align}
where $q(z|x)$ is an auxiliary density induced by a deterministic encoding function $g:\Omega\rightarrow\mathcal{S}^{h-1}$.\footnote{$q(z|x)=\delta(z-g(x))$.} Eq.~(\ref{eq:generative}) provides the basic building block to derive variational bounds on mutual information~\cite{poole2019variational} as well as to obtain several popular contrastive SSL objectives. 

\textbf{$\text{ELBO}_{EBM}$ and variational lower bounds on mutual information}. Our analysis is similar to the one proposed in~\cite{alemi2018fixing}, as relating $\text{ELBO}_{EBM}$ to the mutual information. However, while the work in~\cite{alemi2018fixing} shows that the third line in Eq.~(\ref{eq:generative}) can be expressed as a combination of an upper and a lower bound on the mutual information and it studies its rate-distortion trade-off, our analysis considers only lower bounds to mutual information and it makes an explicit connection to them. Indeed, several contrastive objectives are based on lower bounds on the mutual information between input and latent vectors~\cite{poole2019variational}. We can show that $\text{ELBO}_{EBM}$ is equivalent to these lower bounds under specific conditions for prior $p(z;\Theta)$ and score function $f$.

Specifically, given a uniform prior $p(z;\Theta)$ and $f(x,z)\doteq\log p(x)+\tilde{f}(x,z)$ for all admissible pair $x,z$ and for some arbitrary function $\tilde{f}$, we obtain the following equivalence (see Appendix~\ref{sec:B} for the derivation):
\begin{align}
    \text{ELBO}_{EBM} &= \mathbb{E}_{p(x)}\{\log p(x)\} + \mathbb{E}_{p(x)q(z|x)}\{\tilde{f}(x,z)\} \nonumber\\
    &\quad - \mathbb{E}_{p(x)q(z|x)}\{\log \mathbb{E}_{p(x')}\{e^{\tilde{f}(x',z)}\}\} \nonumber\\
    &= -H(X) + \mathcal{I}_{UBA}(X,Z)
    \label{eq:uba}
\end{align}
where $H(X)=-\mathbb{E}_{p(x)}\{\log p(x)\}$ and $\mathcal{I}_{UBA}(X,Z)$ refers to the popular Unnormalized Barber and Agakov bound on mutual information. Importantly, other well-known bounds can be derived from $\mathcal{I}_{UBA}(X,Z)$ (cf.~\cite{poole2019variational} and Appendix~\ref{sec:B} for further details). Consequently, maximizing $\text{ELBO}_{EBM}$ with respect to the parameters of $f$ (viz. $\theta$) is equivalent to maximize a lower bound on mutual information.

\textbf{$\text{ELBO}_{EBM}$ and InfoNCE~\cite{oord2018representation}}. Now, we are ready to show the derivation of the popular InfoNCE objective. By specifying a non-parametric prior\footnote{The term non-parametric refers to the fact that the samples can be regarded as parameters of the prior and therefore their number increases with the number of samples. Indeed, we have $\Theta=\{\theta;\{x_i\}_{i=1}^n\}$.} $p(z;\Theta)=\frac{\frac{\Gamma(z;\Theta)}{\frac{1}{n}\sum_{k=1}^n e^{f(x_k,z)}}}{\Gamma(\Theta)}$, where $\Gamma(\Theta)=\int\frac{\Gamma(z;\Theta)}{\frac{1}{n}\sum_{k=1}^n e^{f(x_k,z)}}dz$, we achieve the following equality (see Appendix~\ref{sec:C} for the derivation):
\begin{align}
    \text{ELBO}_{EBM} 
    &= E_{\prod_{j=1}^np(x_j,z_j)}\bigg\{\frac{1}{n}\sum_{i=1}\log\frac{e^{f(x_i,z_i)}}{\frac{1}{n}\sum_{k=1}^ne^{f(x_k,z_i)}}\bigg\}  \nonumber\\
    &\qquad- E_{\prod_{j=1}^np(x_j)}\{\log\Gamma(\Theta)\}\nonumber\\
    &\doteq I_{NCE}(X,Z) - E_{\prod_{j=1}^np(x_j)}\{\log\Gamma(\Theta)\}
    \label{eq:nce}
\end{align}
Notably, $\text{ELBO}_{EBM}$ is equivalent to $I_{NCE}(X,Z)$ up to the term $-\log\Gamma(\Theta)$. Maximizing $\text{ELBO}_{EBM}$ has the effect to maximize $I_{NCE}(X,Z)$ and additionally to minimize $\Gamma(\Theta)$, thus ensuring that the prior $p(z;\Theta)$ self-normalizes. However, in practice, people only maximize the InfoNCE objective, disregarding the normalizing term.

{\section{Connection to Unnormalized Barber Agakov Bound}\label{sec:B}}
Firstly, we recall the derivation of the Unnormalized Barber Agakov bound~\cite{poole2019variational} for the mutual information $I_{X,Z}$, adapting it to our notational convention. Secondly, we derive the equivalence relation in Eq.~(\ref{eq:uba}).
\begin{align}
    I_{X,Z} &= \mathbb{E}_{p(x,z)}\bigg\{\log\frac{p(x|z)}{p(x)} \bigg\} \nonumber\\
    &=\mathbb{E}_{p(x)q(z|x)}\bigg\{\log\frac{p(x|z)q(x|z)}{p(x)q(x|z)}\bigg\} \nonumber\\
    &=\mathbb{E}_{p(x)q(z|x)}\bigg\{\log\frac{p(x|z)}{q(x|z)}\bigg\}{+}\mathbb{E}_{p(x)q(z|x)}\bigg\{\log\frac{q(x|z)}{p(x)}\bigg\} \nonumber\\
    &=\mathbb{E}_{q(z)}KL\{p(x|z)\|q(x|z)\}{+} \mathbb{E}_{p(x)q(z|x)}\bigg\{\log\frac{q(x|z)}{p(x)}\bigg\} \nonumber\\
    &\geq \mathbb{E}_{p(x)q(z|x)}\bigg\{\log\frac{q(x|z)}{p(x)}\bigg\} \nonumber\\
    &= \mathbb{E}_{p(x)q(z|x)}\bigg\{\log\frac{p(x)e^{\tilde{f}(x,z)}}{p(x)Z(z)}\bigg\} \nonumber\\
    &= \mathbb{E}_{p(x)q(z|x)}\bigg\{\log\frac{e^{\tilde{f}(x,z)}}{Z(z)}\bigg\} \nonumber\\
    &= \mathbb{E}_{p(x)q(z|x)}\{\tilde{f}(x,z)\}-\mathbb{E}_{p(x)q(z|x)}\{\log Z(z)\} \nonumber\\
    &= \mathbb{E}_{p(x)q(z|x)}\{\tilde{f}(x,z)\}{-}\mathbb{E}_{p(x)q(z|x)}\{\log \mathbb{E}_{p(x')}\{e^{\tilde{f}(x',z)}\}\} \nonumber\\
    &\doteq I_{UBA}(X,Z) \nonumber
\end{align}
where we have introduced both an auxiliary encoder $q(z|x)$ and an auxiliary decoder defined as $q(x|z)=\frac{p(x)e^{\tilde{f}(x,z)}}{Z(z)}$.
Now, we can use the assumptions of a uniform prior $p(z;\Theta)$ and $f(x,z)=\log p(x)+\tilde{f}(x,z)$ to achieve the following inequalities:s
\begin{align}
    \text{ELBO}_{EBM} &= \mathbb{E}_{p(x)q(z|x)}\{f(x,z)\} + \mathbb{E}_{p(x)q(z|x)}\bigg\{\log\frac{p(z;\Theta)}{\Gamma(z;\Theta)}\bigg\} \nonumber \\
    &= \mathbb{E}_{p(x)q(z|x)}\{f(x,z)\} - \mathbb{E}_{p(x)q(z|x)}\{\log\Gamma(z;\Theta)\} \nonumber\\
    &= \mathbb{E}_{p(x)q(z|x)}\{\log p(x)+\tilde{f}(x,z)\} \nonumber\\
    &\qquad- \mathbb{E}_{p(x)q(z|x)}\{\log\mathbb{E}_{p(x')}\{e^{\tilde{f}(x',z)}\} \nonumber\\
    &= \mathbb{E}_{p(x)q(z|x)}\{\log p(x)\}+\mathbb{E}_{p(x)q(z|x)}\{\tilde{f}(x,z)\} \nonumber\\
    &\qquad- \mathbb{E}_{p(x)q(z|x)}\{\log\mathbb{E}_{p(x')}\{e^{\tilde{f}(x',z)}\} \nonumber\\
    &= - H(X) +\mathbb{E}_{p(x)q(z|x)}\{\tilde{f}(x,z)\} \nonumber\\
    &\qquad- \mathbb{E}_{p(x)q(z|x)}\{\log\mathbb{E}_{p(x')}\{e^{\tilde{f}(x',z)}\} \nonumber\\
    &= - H(X) + I_{UBA}(X,Z) \nonumber
\end{align}
\subsection{Other Bounds}
Notably, $I_{UBA}(X,Z)$ cannot be tractably computed due to the computation of $Z(z)$. Therefore, several other bounds have been derived~\cite{poole2019variational} to obtain tractable estimators or optimization objectives, namely:
\begin{enumerate}
    \item $I_{TUBA}(X,Z)$,\footnote{T in TUBA stands for tractable.} which can be used for both optimization and estimation of mutual information (obtained using the inequality $\log s \leq \eta' s - \log\eta' -1$ for all scalar $s,\eta'>0$)
    \begin{align}
        I_{UBA}(X,Z) &= \mathbb{E}_{p(x,z)}\{\tilde{f}(x,z)\}\nonumber\\
        &\qquad-\mathbb{E}_{p(x)q(z|x)}\{\log Z(z)\} \nonumber\\
                     &\geq \mathbb{E}_{p(x,z)}\{\tilde{f}(x,z)\}\nonumber\\
                     &\qquad-\mathbb{E}_{p(x)q(z|x)}\bigg\{\frac{Z(z)}{\eta(z)} +\log \eta(z)-1\bigg\} \nonumber\\
                     &\doteq I_{TUBA}(X,Z) \nonumber
    \end{align}
    \item $I_{NWJ}(X,Z)$, which can be used for both optimization and estimation of mutual information (obtained from $I_{TUBA}(X,Z)$ by imposing $\eta(z)=e$)
    \begin{align}
        I_{TUBA}(X,Z) &= \mathbb{E}_{p(x,z)}\{\tilde{f}(x,z)\}\nonumber\\
        &\qquad-\mathbb{E}_{p(x)q(z|x)}\bigg\{\frac{Z(z)}{\eta(z)} +\log \eta(z)-1\bigg\} \nonumber\\
                      &=\mathbb{E}_{p(x,z)}\{\tilde{f}(x,z)\}\nonumber\\
                      &\qquad-\mathbb{E}_{p(x)q(z|x)}\bigg\{\frac{Z(z)}{e} +1-1\bigg\} \nonumber\\
                      &=\mathbb{E}_{p(x,z)}\{\tilde{f}(x,z)\}-\frac{1}{e}\mathbb{E}_{p(x)q(z|x)}\{Z(z)\} \nonumber\\
                      &=\mathbb{E}_{p(x,z)}\{\tilde{f}(x,z)\}\nonumber\\
                      &\qquad-\mathbb{E}_{p(x')p(x)q(z|x)}\{ e^{\tilde{f}(x',z)-1}\} \nonumber\\
                      &\doteq I_{NWJ}(X,Z) \nonumber
    \end{align}
\end{enumerate} 

{\section{Derivation of InfoNCE}\label{sec:C}}
Consider a prior $p(z)=\frac{\frac{\Gamma(z;\Theta)}{\frac{1}{n}\sum_{k=1}^n e^{f(x_k,z)}}}{\Gamma(\Theta)}$, where $\Gamma(\Theta)=\int\frac{\Gamma(z;\Theta)}{\frac{1}{n}\sum_{k=1}^n e^{f(x_k,z)}}dz$,
\begin{align}
\text{ELBO}_{EBM} & = \mathbb{E}_{p(x)q(z|x)}\{f(x,z)\} - \mathbb{E}_{p(x)q(z|x)}\{\log\Gamma(z;\Theta)\} \nonumber\\
&\qquad+ \mathbb{E}_{p(x)q(z|x)}\{\log p(z;\Theta)\} \nonumber\\ 
&= \mathbb{E}_{p(x_1)q(z|x_1)}\{f(x_1,z)\} \nonumber\\
&\qquad - \mathbb{E}_{p(x_1)q(z|x_1)}\{\log\Gamma(z;\Theta)\} \nonumber\\
&\qquad+ \mathbb{E}_{p(x_1)q(z|x_1)}\{\log p(z;\Theta)\} \nonumber\\
&= \mathbb{E}_{p(x_1)p(x_2)\cdot\cdot\cdot p(x_n)q(z|x_1)}\{f(x_1,z)\} \nonumber\\
&\qquad- \mathbb{E}_{p(x_1)p(x_2)\cdot\cdot\cdot p(x_n)q(z|x_1)}\{\log\Gamma(z;\Theta)\} \nonumber\\
&\qquad + \mathbb{E}_{p(x_1)p(x_2)\cdot\cdot\cdot p(x_n)q(z|x_1)}\{\log p(z;\Theta)\} \nonumber\\
&= \mathbb{E}_{p(x_1)p(x_2)\cdot\cdot\cdot p(x_n)q(z|x_1)}\{f(x_1,z)\} \nonumber\\
&\qquad- \mathbb{E}_{p(x_1)p(x_2)\cdot\cdot\cdot p(x_n)q(z|x_1)}\bigg\{\nonumber\\
&\qquad\qquad\qquad\log\frac{1}{n}\sum_{k=1}^n e^{f(x_k,z)}\bigg\} \nonumber\\
&\qquad -\mathbb{E}_{p(x_1)p(x_2)\cdot\cdot\cdot p(x_n)}\{\log\Gamma(\Theta)\} \nonumber\\
&= \mathbb{E}_{p(x_1)p(x_2)\cdot\cdot\cdot p(x_n)q(z|x_1)}\{\log e^{f(x_1,z)}\} \nonumber\\
&\qquad- \mathbb{E}_{p(x_1)p(x_2)\cdot\cdot\cdot p(x_n)q(z|x_1)}\bigg\{\nonumber\\
&\qquad\qquad\qquad\log\frac{1}{n}\sum_{k=1}^n e^{f(x_k,z)}\bigg\} \nonumber\\
&\qquad-\mathbb{E}_{p(x_1)p(x_2)\cdot\cdot\cdot p(x_n)}\{\log\Gamma(\Theta)\} \nonumber\\
&=
\mathbb{E}_{p(x_1,z)p(x_2)\cdot\cdot\cdot p(x_n)}\bigg\{\log\frac{e^{f(x_1,z)}}{\frac{1}{n}\sum_{k=1}^n e^{f(x_k,z)}}\bigg\}  \nonumber\\
&\qquad- \mathbb{E}_{p(x_1)p(x_2)\cdot\cdot\cdot p(x_n)}\{\log\Gamma(\Theta)\}\nonumber\\
&=
\mathbb{E}_{p(x_1,z_1)p(x_2,z_2)\cdot\cdot\cdot p(x_n,z_K)}\bigg\{\nonumber\\
&\qquad\qquad\qquad\log\frac{e^{f(x_1,z_1)}}{\frac{1}{n}\sum_{k=1}^n e^{f(x_k,z_1)}}\bigg\}  \nonumber\\
&\qquad- \mathbb{E}_{p(x_1)p(x_2)\cdot\cdot\cdot p(x_n)}\{\log\Gamma(\Theta)\} \nonumber\\
&=
\frac{1}{n}\sum_{i=1}\mathbb{E}_{p(x_1,z_1)p(x_2,z_2)\cdot\cdot\cdot p(x_n,z_K)}\bigg\{\nonumber\\
&\qquad\qquad\qquad\log\frac{e^{f(x_i,z_i)}}{\frac{1}{n}\sum_{k=1}^ne^{f(x_k,z_i)}}\bigg\}  \nonumber\\
&\qquad- \mathbb{E}_{p(x_1)p(x_2)\cdot\cdot\cdot p(x_n)}\{\log\Gamma(\Theta)\} \nonumber\\
&=
\mathbb{E}_{p(x_1,z_1)p(x_2,z_2)\cdot\cdot\cdot p(x_n,z_K)}\bigg\{\nonumber\\
&\qquad\qquad\qquad\frac{1}{n}\sum_{i=1}\log\frac{e^{f(x_i,z_i)}}{\frac{1}{n}\sum_{k=1}^ne^{f(x_k,z_i)}}\bigg\}  \nonumber\\
&\qquad- \mathbb{E}_{p(x_1)p(x_2)\cdot\cdot\cdot p(x_n)}\{\log\Gamma(\Theta)\} \nonumber\\
&\doteq I_{NCE}(X,Z) - \mathbb{E}_{\prod_{j=1}^np(x_j)}\{\log\Gamma(\Theta)\} \nonumber
\end{align}

{\section{Derivation of ProtoCPC: A Lower Bound of InfoNCE}\label{sec:D}}
ProtoCPC~\cite{lee2022prototypical} can be viewed as a  lower bound of the InfoNCE objective. To save space, we use notation $\mathbb{E}$ to refer to $\mathbb{E}_{\prod_{j=1}^np(x_j,z_j)}$. Therefore, we have that
\begin{align}
    I_{NCE} &= \mathbb{E}\bigg\{\frac{1}{n}\sum_{i=1}\log\frac{e^{f(x_i,z_i)}}{\frac{1}{n}\sum_{k=1}^ne^{f(x_k,z_i)}}\bigg\} \nonumber\\
    &= \mathbb{E}\bigg\{\frac{1}{n}\sum_{i=1}\log\frac{e^{\sum_{c=1}^Cp_c^t(x_i)\log p_c^s(x_i)}}{\frac{1}{n}\sum_{k=1}^ne^{\sum_{c=1}^Cp_c^t(x_k)\log p_c^s(x_i)}}\bigg\} \nonumber\\
    &= \mathbb{E}\left\{\frac{1}{n}\sum_{i=1}\log\frac{\splitfrac{e^{\sum_{c=1}^C p_c^t(x_i)\log e^{z_{c,x_i}^s/\tau_s}}\cdot} {e^{-\sum_{c=1}^C p_c^t(x_i)\log \sum_{c'=1}^C e^{z_{c',x_i}^s/\tau_s}}}}{\frac{1}{n}\sum_{k=1}^ne^{\sum_{c=1}^Cp_c^t(x_k)\log p_c^s(x_i)}}\right\} \nonumber\\
    &= \mathbb{E}\left\{\frac{1}{n}\sum_{i=1}\log\frac{\splitfrac{e^{\sum_{c=1}^C p_c^t(x_i)\log e^{z_{c,x_i}^s/\tau_s}}\cdot} {e^{-\sum_{c=1}^C p_c^t(x_i)\log \sum_{c'=1}^C e^{z_{c',x_i}^s/\tau_s}}}}{\splitfrac{\frac{1}{n}\sum_{k=1}^ne^{\sum_{c=1}^C p_c^t(x_k)\log e^{z_{c,x_i}^s/\tau_s}}\cdot} {e^{-\sum_{c=1}^C p_c^t(x_k)\log \sum_{c'=1}^C e^{z_{c',x_i}^s/\tau_s}}}}\right\} \nonumber\\
    &= \mathbb{E}\left\{\frac{1}{n}\sum_{i=1}\log\frac{\splitfrac{e^{\sum_{c=1}^C p_c^t(x_i)\log e^{z_{c,x_i}^s/\tau_s}}\cdot}{e^{-\log \sum_{c'=1}^C e^{z_{c',x_i}^s/\tau_s}}}}{\splitfrac{\frac{1}{n}\sum_{k=1}^ne^{\sum_{c=1}^C p_c^t(x_k)\log e^{z_{c,x_i}^s/\tau_s}}\cdot}{e^{-\log \sum_{c'=1}^C e^{z_{c',x_i}^s/\tau_s}}}}\right\} \nonumber\\
    &= \mathbb{E}\bigg\{\frac{1}{n}\sum_{i=1}\log\frac{e^{\sum_{c=1}^C p_c^t(x_i)\log e^{z_{c,x_i}^s/\tau_s}}}{\frac{1}{n}\sum_{k=1}^ne^{\sum_{c=1}^C p_c^t(x_k)\log e^{z_{c,x_i}^s/\tau_s}}}\bigg\} \nonumber\\
    &= \mathbb{E}\bigg\{\frac{1}{n}\sum_{i=1}\log\frac{e^{\sum_{c=1}^C p_c^t(x_i)z_{c,x_i}^s/\tau_s}}{\frac{1}{n}\sum_{k=1}^ne^{\sum_{c=1}^C p_c^t(x_k)z_{c,x_i}^s/\tau_s}}\bigg\} \nonumber\\
    &\geq \mathbb{E}\bigg\{\frac{1}{n}\sum_{i=1}\log\frac{e^{\sum_{c=1}^C p_c^t(x_i)z_{c,x_i}^s/\tau_s}}{\frac{1}{n}\sum_{k=1}^n\sum_{c=1}^C p_c^t(x_k)e^{z_{c,x_i}^s/\tau_s}}\bigg\} \nonumber\\
    &= \mathbb{E}\bigg\{\frac{1}{n}\sum_{i=1}\log\frac{e^{\sum_{c=1}^C p_c^t(x_i)z_{c,x_i}^s/\tau_s}}{\sum_{c=1}^C \frac{1}{n}\sum_{k=1}^n p_c^t(x_k)e^{z_{c,x_i}^s/\tau_s}}\bigg\} \nonumber\\
    &= \mathbb{E}\bigg\{\frac{1}{n}\sum_{i=1}\log\frac{e^{\sum_{c=1}^C p_c^t(x_i)z_{c,x_i}^s/\tau_s}}{\sum_{c=1}^C q_c^t e^{z_{c,x_i}^s/\tau_s}}\bigg\} \nonumber\\
    &= \mathbb{E}\bigg\{\frac{1}{n}\sum_{i=1}\log\frac{e^{ p^t(x_i)^T z_{x_i}^s/\tau_s}}{\sum_{c=1}^C q_c^t e^{z_{c,x_i}^s/\tau_s}}\bigg\} \doteq I_{ProtoCPC} \nonumber
\end{align}
Specifically, the second equality comes from the fact that $f(x,z)=\sum_{c=1}^C p_c^t(x)\log p_c^s(x)$, where $s,t$ stand for student and teacher networks, respectively, and $p_c^\cdot(x)$ is the $c$-th entry of the vector obtained by applying a softmax on the embedding of the corresponding network. The inequality in the derivation is obtained by applying Jensen's inequality to the denominator. Finally, $q_c^t=\frac{1}{n}\sum_{k=1}^n p_c^t(x_k)$ corresponds to the prototype for class $c$.

{\section{Details about Negative-Free Methods}\label{sec:E}}
We can specify different definitions for $\mathcal{L}_{NF}(\Theta)$, namely:
\begin{enumerate}
    \item \textbf{Barlow Twins.} The approach enforces the cross-correlation matrix to be close to the identity matrix:
    $$\mathcal{L}_{NF}(\Theta)=-\|CCorr(G,G')\odot\Lambda - I\|_F^2$$
    where $\|\cdot\|_F$ is the Frobenius norm, $\Lambda=J\lambda + (1-\lambda)I$, $I$ is the identity matrix, $\lambda$ is a positive hyperparameter, $J$ is a matrix of ones,
    $CCorr(G,G')=H^TH'$ is the sample cross-correlation matrix, $G=[g(x_1),\dots,g(x_n')]^T\in\mathbb{R}^{n\times h}$ and $G'=[g(x_1'),\dots,g(x_n')]^T\in\mathbb{R}^{n\times h}$ and
    $$g(x_i)=\text{BN}(\text{Proj}(enc(x_i)))$$
    Note that $BN$ is a batch normalization layer, $Proj$ is a projection head usually implemented using a multi-layer perceptron and $enc$ is the encoder.
    \item \textbf{W-MSE.} This approach is similar to Barlow Twins. The main difference lie in the fact that sample cross-correlation matrix is enforced to be an identity matrix thanks to a whitening layer:
    $$\mathcal{L}_{NF}(\Theta)=-Tr(G\otimes G')$$
    where $Tr(G\otimes G')$ computes the trace of the outer product for matrices $G=[g(x_1),\dots,g(x_n')]^T\in\mathbb{R}^{n\times h}$ and $G'=[g(x_1'),\dots,g(x_n')]^T\in\mathbb{R}^{n\times h}$.
    $$g(x_i)=\text{L2-Norm}(\text{Whitening}(\text{Proj}(enc(x_i))))$$
    $\text{L2-Norm}$ is a normalization layer based on $L_2$ norm.
    \item \textbf{VICReg.} This approach attempts to simplify the architecture of Barlow Twins by introducing an invariance regularization term in the score function, thus avoiding to use a batch normalization layer:
    \begin{align}
        \mathcal{L}_{NF}(\Theta)&=-\lambda Tr((G-G')\otimes(G-G'))\nonumber\\
        &-\mu[v(G)+v(G')]-\nu[w(G)+w(G')] \nonumber
    \end{align}
    where $\lambda, \mu, \nu$ are positive hyperparameters and $G=[g(x_1),\dots,g(x_n')]^T\in\mathbb{R}^{n\times h}$ and $G'=[g(x_1'),\dots,g(x_n')]^T\in\mathbb{R}^{n\times h}$. The first addend enforces the representation to be invariant to the data augmentation, whereas the other two addends enforce the sample covariance matrix to be diagonal. Indeed, the second addend forces the diagonal elements of the sample covariance matrices to be unitary, namely:
    $$v(G)=\sum_{j=1}^h\max\{0,1-\sqrt{Var\{G_{:j}\}+\epsilon}\}$$
    where $\epsilon>0$ is used to avoid numerical issues, $Var$ computes the variance for each column of matrix $G$. The third addend ensures that the off-diagonal elements of the sample covariance matrix approach to zero:
    $$c(G)=\|Cov(G)\odot(J-I)\|_F^2$$
    with $Cov(H)=G^TG$. Therefore, these last two addends have a similar behaviour to the score function of Barlow Twins.
    Finally, the resulting function is simplified.
    $$g(x_i)=\text{Proj}(enc(x_i))$$
\end{enumerate}

\begin{figure*}
    \centering
    \includegraphics[width=0.7\textwidth]{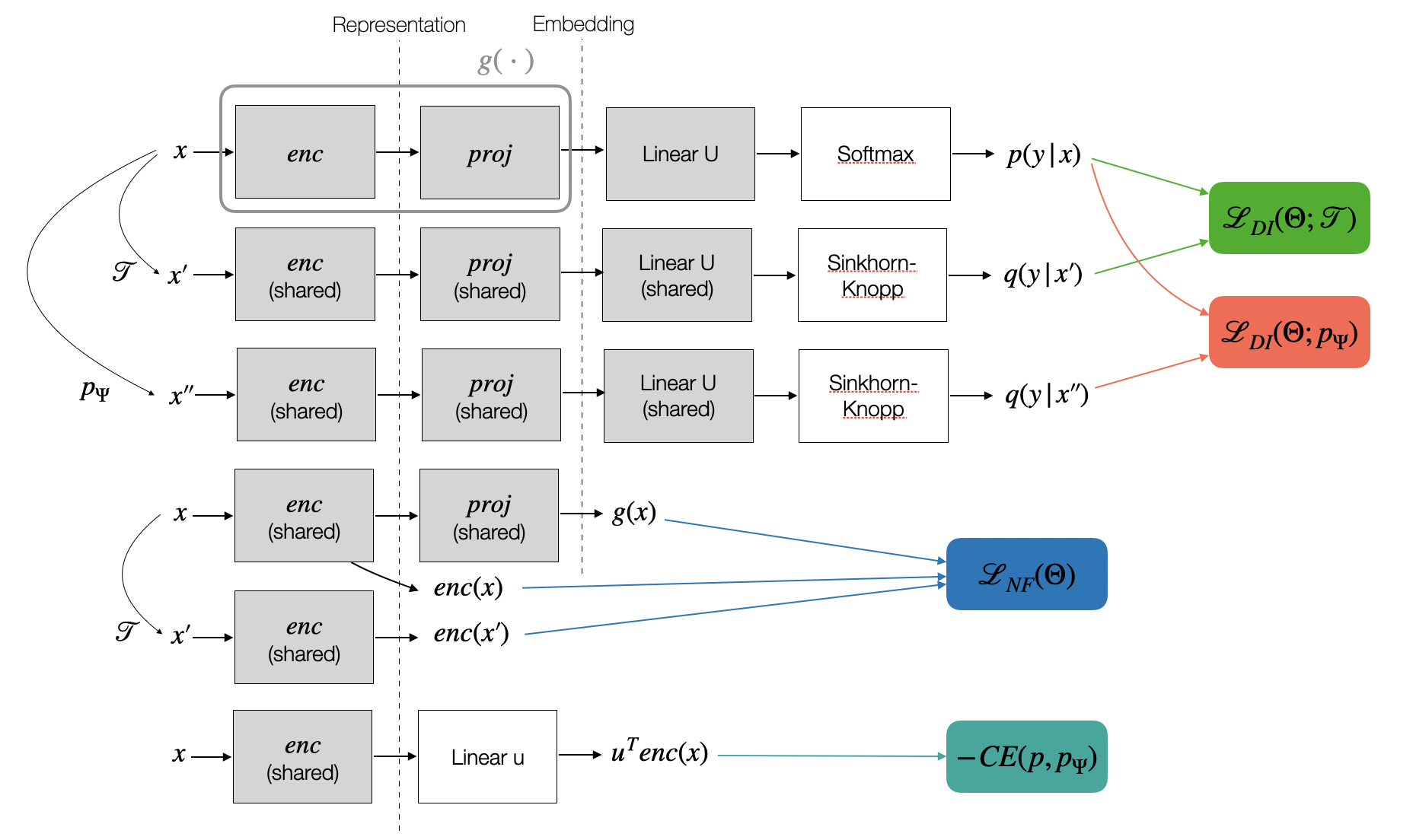}
    \caption{Diagram of the whole model. Grey boxes are shared among different rows. $proj^*$} has the same architecture of $proj$, but different weights and no $L_2$ normalization layer in its output.
     \label{fig:model}
\end{figure*}

{\section{Whole Model and Training Algorithm}\label{sec:training}}
Figure~\ref{fig:model} shows the whole GEDI architecture and how to compute the different losses.

\begin{figure}
     \centering
     \begin{subfigure}[b]{0.33\linewidth}
         \centering
         \includegraphics[width=0.9\textwidth]{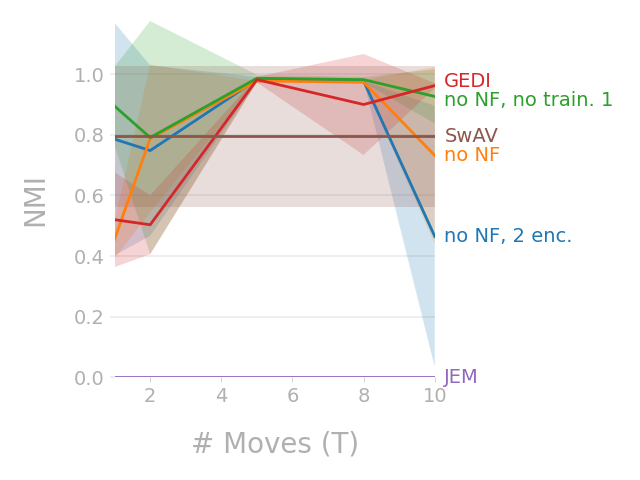}
         \caption{$c=2$}
     \end{subfigure}%
     \begin{subfigure}[b]{0.33\linewidth}
         \centering
         \includegraphics[width=0.9\textwidth]{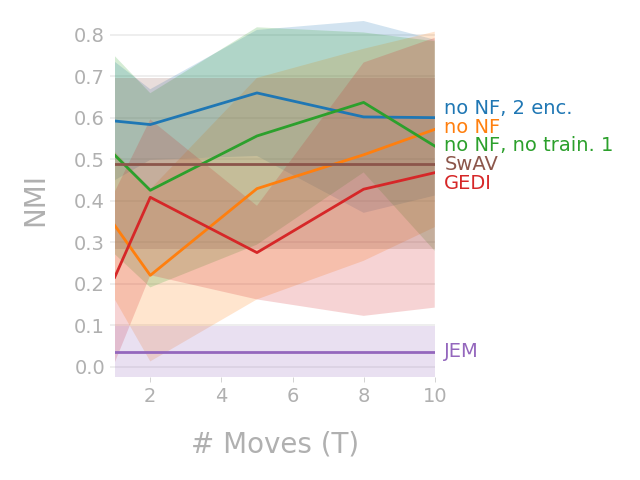}
         \caption{$c=4$}
     \end{subfigure}%
     \begin{subfigure}[b]{0.33\linewidth}
         \centering
         \includegraphics[width=0.9\textwidth]{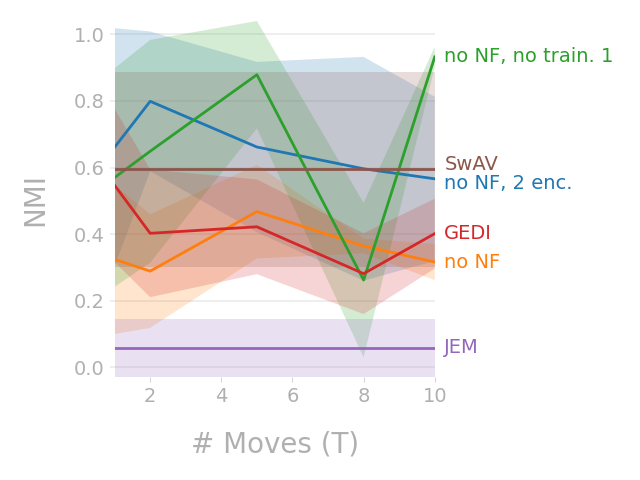}
         \caption{$c=6$}
     \end{subfigure}%
     \\
     \begin{subfigure}[b]{0.33\linewidth}
         \centering
         \includegraphics[width=0.9\textwidth]{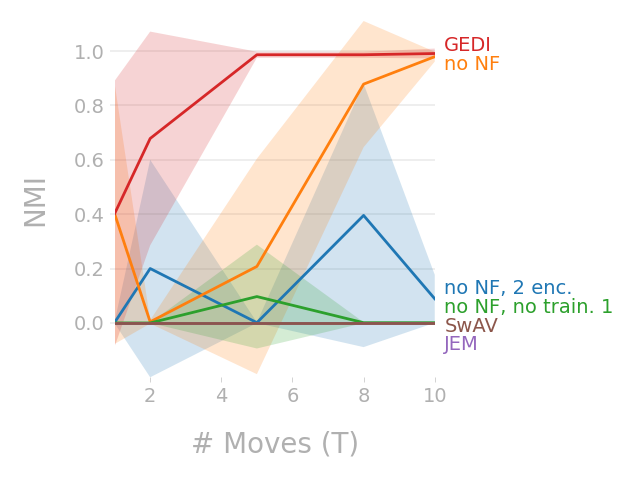}
         \caption{$c=2$}
     \end{subfigure}%
     \begin{subfigure}[b]{0.33\linewidth}
         \centering
         \includegraphics[width=0.9\textwidth]{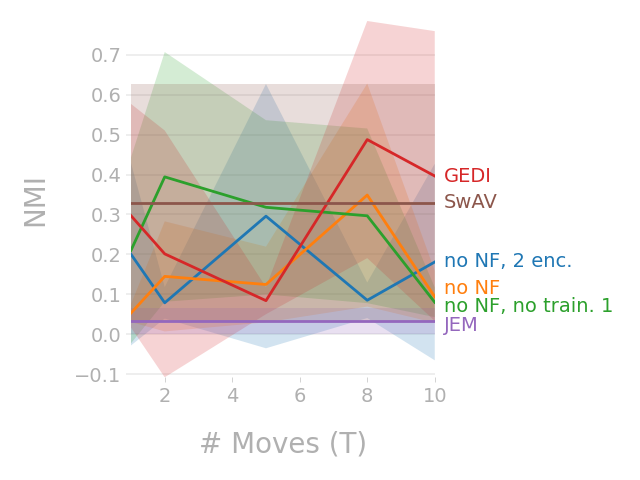}
         \caption{$c=4$}
     \end{subfigure}%
     \begin{subfigure}[b]{0.33\linewidth}
         \centering
         \includegraphics[width=0.9\textwidth]{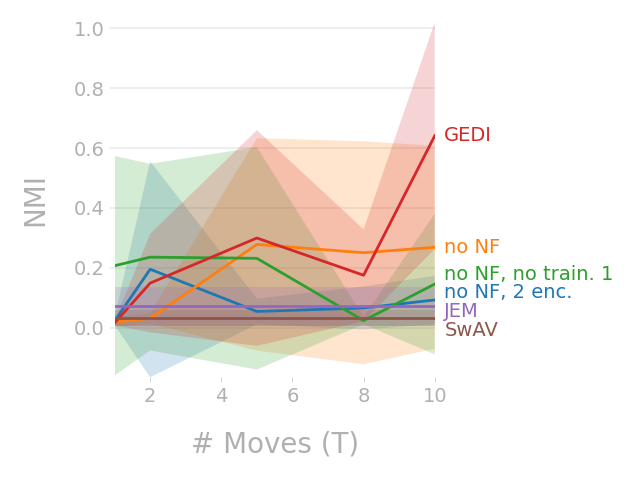}
         \caption{$c=6$}
     \end{subfigure}%
     \caption{Additional experiments on synthetic data, viz. (a-c) on moons, (d-f) on circles. The curves are obtained by choosing different values of $T$ for DAM, namely $T\in\{1,2,5,8,10\}$.}
     \label{fig:classes_toy}
\end{figure}

{\section{Hyperparameters for Synthetic Data}\label{sec:hyperparams_synth}}
For the backbone $enc$, we use a MLP with two hidden layers and 100 neurons per layer, an output layer with 2 neurons and LeakyReLU activation functions. For the projection head $proj$, we use a MLP with one hidden layer and 4 and an output layer with 2 neurons (batch normalization is used in all layers) and final $L_2$ normalization.
For JEM, we use an output layer with only one neuron. All methods use a batch size of 400.
Baseline JEM (following the original paper):
\begin{itemize}
    \item Number of iterations $100K$
    \item Learning rate $1e-3$
    \item Optimizer Adam $\beta_1=0$, $\beta_2=0.9$
    \item SGLD steps $1$
    \item Buffer size 10000
    \item Reinitialization frequency $0.05$
    \item SGLD step-size $1$
    \item SGLD noise $0.01$
\end{itemize}
And for self-supervised learning methods, please refer to Table~\ref{tab:hyperparams}.
\begin{table*}[ht]

  \caption{Hyperparameters (in terms of sampling, optimizer, objective and data augmentation) used in all experiments.}
  \label{tab:hyperparams}
  \centering
\begin{tabular}{@{}llrrrr@{}}
\toprule
\textbf{Class} & \textbf{Name param.} & \textbf{SVHN} & \textbf{CIFAR-10} & \textbf{CIFAR-100} & \textbf{Addition} \\
\midrule
\multirow{5}{*}{Data augment.} & Color jitter prob. & 0.1 & 0.1 & 0.1 & 0.1 \\
& Gray scale prob. & 0.1 & 0.1 & 0.1 & 0.1 \\
& Random crop & Yes & Yes & Yes & Yes \\
& Additive Gauss. noise (std) & 0.03 & 0.03 & 0.03 & 0.3 \\
& Random horizontal flip & No & Yes & Yes & No \\
\midrule
            \multicolumn{6}{c}{Training 1} \\
\midrule
\multirow{5}{*}{SGLD} & SGLD iters & 20 & 20 & 20 & 10 \\
& Buffer size & 10k & 10k & 10k & 10k \\
& Reinit. frequency & 0.05 & 0.05 &0.05 & 0.05 \\
& SGLD step-size & 1 & 1 & 1 & 1 \\
& SGLD noise & 0.01 & 0.01 & 0.01 & 0.01 \\
\midrule
\multirow{6}{*}{Optimizer} & Batch size & 64 & 64 & 64 & 60 \\
& $\text{Iters}_1$ & 100k & 70k & 70k & Sec.~\ref{sec:hyperparams_nesy}\\
& Adam $\beta_1$ & 0.9 & 0.9 & 0.9 & 0.9 \\
& Adam $\beta_2$ & 0.999 & 0.999 & 0.999 & 0.999 \\
& Learning rate & $1e-4$ & $1e-4$ & $1e-4$ & $1e-4$ \\
\midrule
            \multicolumn{6}{c}{Training 2} \\
\midrule
\multirow{5}{*}{SGLD} & SGLD iters & idem & idem & idem & idem \\
& Buffer size & idem & idem & idem & idem \\
& Reinit. frequency & idem & idem & idem & idem \\
& SGLD step-size & idem & idem & idem & idem \\
& SGLD noise & idem & idem & idem & idem \\
\midrule
\multirow{6}{*}{Optimizer} & Batch size & idem & idem & idem & idem \\
& $\text{Iters}_2$ & idem & idem & idem & idem \\
& Adam $\beta_1$ & idem & idem & idem & idem \\
& Adam $\beta_2$ & idem & idem & idem & idem \\
& Learning rate & idem & idem & idem & idem \\
\midrule
\multirow{2}{*}{DAM} & $\epsilon$ & 0.03 & 0.03 & 0.03 & 0.03 \\
& $T$ & 10 & 10 & 10 & 10 \\
\midrule
\multirow{5}{*}{Weights for losses} & $-CE(p,p_\Psi)$ & 1 & 1 & 1 & 1 \\
&$\mathcal{L}_{NF}(\Theta)$ & 1/batch & 1/batch & 1/batch & 1/batch \\
& $\mathcal{L}_{DI}(\Theta,\mathcal{T})$ & 1000 & 1000 & 1000 & 1000 \\
& $\mathcal{L}_{DI}(\Theta,p_\Psi)$ & 500 & 500 & 500 & 500 \\
& $\mathcal{L}_{NeSY}(\Theta)$ & - & - & - & 0/3000 \\
\bottomrule
\end{tabular}
\end{table*}
{\section{Additional Experiments on Synthetic Data}\label{sec:synth}}
We run additional experiments on synthetic data changing the value of $c=\{2,4,6\}$. Results are shown in Figure~\ref{fig:classes_toy}. When $c=4$ or $c=6$, we observe that DAM almost never improves the performance and this is quite reasonable as we are forcing connected clusters to be clustered by disjoint ones (thus breaking the underlying manifold assumption about connectedness).

{\section{Hyperparameters for SVHN, CIFAR-10, CIFAR-100}\label{sec:hyperparams_real}}
\begin{table}[ht]
  \caption{Resnet architecture. Conv2D(A,B,C) applies a 2d convolution to input with B channels and produces an output with C channels using stride (1, 1), padding (1, 1) and kernel size (A, A).}
  \label{tab:backbone}
  \centering
\begin{tabular}{@{}lrr@{}}
\toprule
\textbf{Name} & \textbf{Layer} & \textbf{Res. Layer} \\
\midrule
\multirow{6}{*}{Block 1} & Conv2D(3,3,F) & \multirow{2}{*}{AvgPool2D(2)} \\
& LeakyRELU(0.2) & \\
& Conv2D(3,F,F) & \multirow{2}{*}{Conv2D(1,3,F) no padding}\\
& AvgPool2D(2) & \\
& \cline{1-2}\\
& \multicolumn{2}{c}{Sum} \\
\midrule
\multirow{5}{*}{Block 2} & LeakyRELU(0.2) & \\
& Conv2D(3,F,F) & \\
& LeakyRELU(0.2) & \\
& Conv2D(3,F,F) & \\
& AvgPool2D(2) & \\
\midrule
\midrule
\multirow{4}{*}{Block 3} & LeakyRELU(0.2) & \\
& Conv2D(3,F,F) & \\
& LeakyRELU(0.2) & \\
& Conv2D(3,F,F) & \\
\midrule
\multirow{5}{*}{Block 4} & LeakyRELU(0.2) & \\
& Conv2D(3,F,F) & \\
& LeakyRELU(0.2) & \\
& Conv2D(3,F,F) & \\
& AvgPool2D(all) & \\
\bottomrule
\end{tabular}
\end{table}
For the backbone $enc$, we use a ResNet with 8 layers as in~\cite{duvenaud2021no}, where its architecture is shown in Table~\ref{tab:backbone}. For the projection head $proj$, we use a MLP with one hidden layer and $2*F$ neurons and an output layer with $F$ neurons (batch normalization is used in all layers) and final $L_2$ normalization. $F=128$ for SVHN, CIFAR-10 (1 million parameters) and $F=256$ for CIFAR-100 (4.1 million parameters).
 For JEM, we use the same settings of~\cite{duvenaud2021no}. All methods use a batch size of 64.
Baseline JEM (following the original paper):
\begin{itemize}
    \item Number of epochs $100$
    \item Learning rate $1e-4$
    \item Optimizer Adam
    \item SGLD steps $20$
    \item Buffer size 10000
    \item Reinitialization frequency $0.05$
    \item SGLD step-size $1$
    \item SGLD noise $0.01$
    \item Data augmentation (Gaussian noise) $0.03$
\end{itemize}
And for self-supervised learning methods, please refer to Table~\ref{tab:hyperparams}.

\begin{table*}
  \caption{Supervised linear evaluation in terms of accuracy on test set (SVHN, CIFAR-10, CIFAR-100). The linear classifier is trained for 100 epochs using SGD with momentum, learning rate $1e-3$ and batch size 100.}
  \label{tab:acc}
  \centering
  \begin{tabular}{@{}lllllll@{}}
    \toprule
    \textbf{Dataset} & \textbf{JEM} & \textbf{Barlow} & \textbf{SwAV} & \textbf{No NF, 2 enc.} & \textbf{No NF} & \textbf{GEDI} \\
    \midrule
    SVHN & 0.20 & 0.84 & 0.44 & 0.29 & 0.52 & 0.56 \\
    CIFAR-10 & 0.23 & 0.63 & 0.53 & 0.48 & 0.50 & 0.51 \\
    CIFAR-100 & 0.03 & 0.35 & 0.14 & 0.12 & 0.15 & 0.15 \\
    \bottomrule
  \end{tabular}
\end{table*}

\begin{table*}
  \caption{Generative performance in terms of Frechet Inception Distance (FID) (SVHN, CIFAR-10, CIFAR-100). The lower the values the better the performance are.}
  \label{tab:FID}
  \centering
  \begin{tabular}{@{}lllllll@{}}
    \toprule
    \textbf{Dataset} & \textbf{JEM} & \textbf{Barlow} & \textbf{SwAV} & \textbf{No NF, 2 enc.} & \textbf{No NF} & \textbf{GEDI} \\
    \midrule
    SVHN & 166 & 454 & 489 & \textbf{158} & 173 & 208 \\
    CIFAR-10 & 250 & 413 & 430 & \textbf{209} & 265 & 236 \\
    CIFAR-100 & 240 & 374 & 399 & \textbf{210} & 237 & 244 \\
    \bottomrule
  \end{tabular}
\end{table*}

\begin{table*}
  \caption{OOD detection in terms of AUROC on test set (CIFAR-10, CIFAR-100). Training is performed on SVHN.}
  \label{tab:oodsvhn}
  \centering
  \begin{tabular}{@{}lllllll@{}}
        \toprule
        \textbf{Dataset} & \textbf{JEM} & \textbf{Barlow} & \textbf{SwAV} & \textbf{No NF, 2 enc.} & \textbf{No NF} & \textbf{GEDI} \\
    \midrule
    CIFAR-10 & 0.75 & 0.43 & 0.21 & 0.76 & \textbf{0.97} & 0.94 \\
    CIFAR-100 & 0.75 & 0.5 & 0.28 & 0.75 & \textbf{0.94} & \textbf{0.93} \\
    \bottomrule
  \end{tabular}
\end{table*}

\begin{table*}
  \caption{OOD detection in terms of AUROC on test set (SVHN, CIFAR-100). Training is performed on CIFAR-10.}
  \label{tab:oodCIFAR-10}
  \centering
  \begin{tabular}{@{}lllllll@{}}
        \toprule
        \textbf{Dataset} & \textbf{JEM} & \textbf{Barlow} & \textbf{SwAV} & \textbf{No NF, 2 enc.} & \textbf{No NF} & \textbf{GEDI} \\
    \midrule
    SVHN & \textbf{0.43} & 0.31 & 0.24 & \textbf{0.43} & 0.31 & 0.31 \\
    CIFAR-100 & 0.54 & \textbf{0.56} & 0.51 & 0.53 & 0.53 & \textbf{0.55} \\
    \bottomrule
  \end{tabular}
\end{table*}

\begin{table*}
  \caption{OOD detection in terms of AUROC on test set (SVHN, CIFAR-10). Training is performed on CIFAR-100.}
  \label{tab:oodCIFAR-100}
  \centering
  \begin{tabular}{@{}lllllll@{}}
        \toprule
        \textbf{Dataset} & \textbf{JEM} & \textbf{Barlow} & \textbf{SwAV} & \textbf{No NF, 2 enc.} & \textbf{No NF} & \textbf{GEDI} \\
    \midrule
    SVHN & 0.48 & 0.43 & \textbf{0.50} & 0.47 & 0.32 & 0.38 \\
    CIFAR-10 & 0.48 & 0.42 & 0.46 & 0.47 & \textbf{0.49} & \textbf{0.50} \\
    \bottomrule
  \end{tabular}
\end{table*}

{\section{Additional Experiments on SVHN, CIFAR-10, CIFAR-100}}
We conduct a linear probe evaluation of the representations learnt by the different models Table~\ref{tab:acc}. These experiments provide insights on the capabilities of learning representations producing linearly separable classes. From Table~\ref{tab:acc}, we observe a large difference in results between Barlow and SwAV. Our approach provides interpolating results between the two approaches.

We also evaluate the generative performance in terms of Frechet Inception Distance (FID). From Table~\ref{tab:FID}, we observe that GEDI outperforms all self-supervised baselines by a large margin, achieving comparable performance to JEM.

Additionally, we evaluate the performance in terms of OOD detection, by following the same methodology used in~\cite{grathwohl2020your}. We use the OOD score criterion proposed in~\cite{grathwohl2020your}, namely $s(x)=-\|\frac{\partial\log p_\Psi(x)}{\partial x}\|_2$. From Table~\ref{tab:oodsvhn}, we observe that GEDI achieves almost optimal performance. While these results are exciting, it is important to mention that they are not generally valid. Indeed, when training on CIFAR-10 and performing OOD evaluation on the other datasets, we observe that all approaches achieve similar performance both on CIFAR-100 and SVHN, suggesting that all datasets are considered in-distribution, see Table~\ref{tab:oodCIFAR-10}. A similar observation is obtained when training on CIFAR-100 and evaluating on CIFAR-10 and SVHN, see Table~\ref{tab:oodCIFAR-100}. Importantly, this is a phenomenon which has been only recently observed by the scientific community on generative models. Tackling this problem is currently out of the scope of this work. For further discussion about the issue, we point the reader to the works in~\cite{nalisnick2019deep}.

{\section{Details on the MNIST addition experiment.}\label{sec:hyperparams_nesy}}
We now discuss the details on the MNIST addition experiment.

\subsection{Hyperparameters}
For the backbone $enc$, we use a ResNet with 8 layers as in~\cite{duvenaud2021no}, where its architecture is shown in Table~\ref{tab:backbone}. For the projection head $proj$, we use a MLP with one hidden layer and 256 neurons and an output layer with 128 neurons (batch normalization is used in all layers) and final $L_2$ normalization. 
The number of epochs for both training phases for the three settings, i.e. $100$ examples, $1 000$ examples and $10 000$ examples are $100$, $30$ and $5$ epochs respectively. These were selected by the point at which the loss curves flatten out.

\subsection{Data generation}
The data was generated by uniformly sampling pairs $a,b$ such that $0 \le a \le 9$, $0 \le b \le 9$ and $0 \le a+b \le 9$. For each triplet $(a,b,c)$, we assigned to $a,b,c,$ random MNIST images with corresponding labels, without replacement.

\subsection{Calculating the constraint}
To calculate the constraint, we group the three images of each triplet consecutively in the batches, hence why the batch size is a multiple of 3. To calculate the probability of the constraint, we used an arithmetic circuit compiled from the DeepProbLog program that implements this constraint ~\cite{manhaeve2018deepproblog}.

\end{document}